\documentclass[letterpaper]{article}
\usepackage{aaai18}
\usepackage{times}
\usepackage{helvet}
\usepackage{courier}
\usepackage{url}
\usepackage{graphicx}
\usepackage[utf8]{inputenc} 
\usepackage[T1]{fontenc}    
\usepackage{url}            
\usepackage{booktabs}       
\usepackage{amsfonts}       
\usepackage{nicefrac}       
\usepackage{microtype}      
\usepackage{epsf, setspace, latexsym}
\usepackage{xspace, amsmath, amssymb, bm}
\usepackage{color,amsthm}
\usepackage{graphicx}
\usepackage{color}
\usepackage{subcaption}
\usepackage{graphicx}
\usepackage{algorithm2e}

\def\QED{\mbox{\rule[0pt]{1.5ex}{1.5ex}}}
\def\endproof{\hspace*{\fill}\tilde \QED\par\endtrivlist\unskip}

\def\bfe{{{\mathbf e}}}

\def\diag{\ensuremath{\mathrm{diag}}}
\def\A{\ensuremath{{\mathcal A}}\xspace}

\def\O{\ensuremath{{\mathcal O}}\xspace}
\def\R{\ensuremath{{\mathbb R}}\xspace}

\def\sign{\ensuremath{\mathrm{sgn}}}
\DeclareMathOperator*{\argmin}{argmin}
\DeclareMathOperator*{\argmax}{argmax}

 \def\HAPPY{DUTI\xspace}
\title{Training Set Debugging Using Trusted Items}
\author{Xuezhou Zhang \and Xiaojin Zhu \and Stephen Wright\\
		\{zhangxz1123, jerryzhu, swright\}@cs.wisc.edu\\
	Department of Computer Sciences, University of Wisconsin-Madison\\
}
\begin{document}
%
\maketitle

\begin{abstract}
Training set bugs are flaws in the data that adversely
affect machine learning. The training set is usually too large for
manual inspection, but one may have the resources to verify a few
trusted items. The set of trusted items may not by itself be adequate
for learning, so we propose an algorithm that uses these items to
identify bugs in the training set and thus improves learning. Specifically, our approach seeks
the smallest set of changes to the training set labels such that the
model learned from this corrected training set predicts labels of the
trusted items correctly. We flag the items whose labels are changed
as potential bugs, whose labels can be checked for veracity by human
experts. To find the bugs in this way is a challenging combinatorial bilevel optimization
problem, but it can be relaxed into a continuous optimization problem.
Experiments on
toy and real data demonstrate that our approach can identify
training set bugs effectively and suggest appropriate changes to the
labels. Our algorithm is a step toward trustworthy machine learning.
\end{abstract}

\section{Introduction}
A good training set is essential for machine learning. The presence of bugs -- mislabeled training items\footnote{We focus on label bugs for simplicity, though our framework can be extended to feature bugs in a straightforward manner.} -- has adverse effects on learning~\cite{brodley1999,guruswami2009hardness,caramanis2008learning}. 
Bugs can appear as outliers that are relatively easy to detect, or as systematic biases.
Systematic bugs are much harder to detect because the data appear self-consistent. 

We propose a novel algorithm \HAPPY (Debugging Using Trusted Items) which can detect both outlier and systematic training set bugs. 
In addition, it can propose fixes, namely the corrected label for the bugs. 
To do so, \HAPPY utilizes the knowledge of the machine learning algorithm and a small set of additional ``trusted items''.
At its core, \HAPPY finds the smallest changes to the training set such that, when trained on the changed training set, the learned model agrees with the trusted items.  The changes are then shown to a domain expert as suggested bug fixes.
We will show how \HAPPY can be relaxed and solved efficiently using continuous optimization, and we demonstrate its debugging efficacy on multiple data sets.
All code and data are published at \url{http://pages.cs.wisc.edu/~jerryzhu/DUTI}.
\begin{figure}[t!]
	\centering
	\begin{minipage}[t]{0.45\columnwidth}
		\centering
		\includegraphics[width=.9\columnwidth]{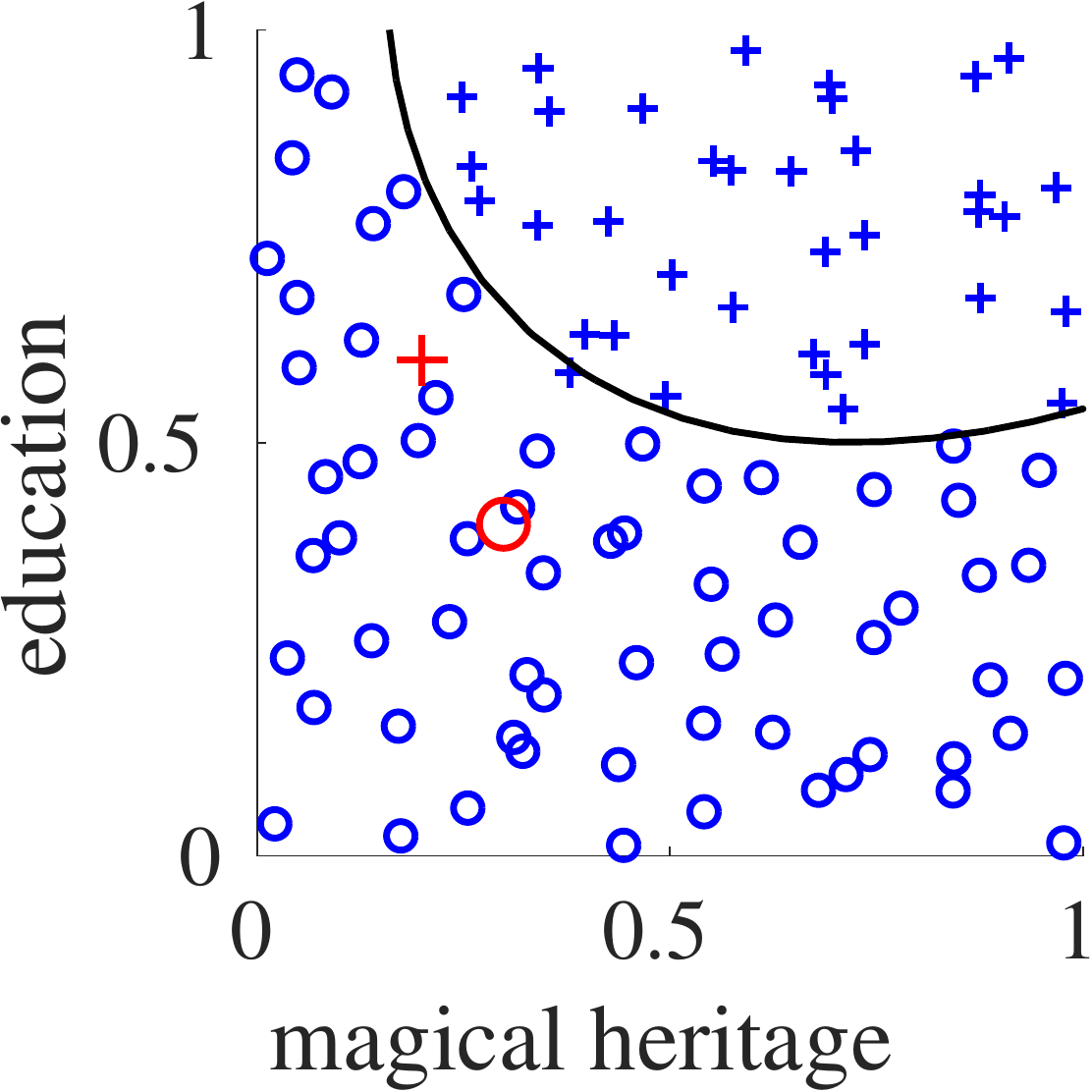} 
		\subcaption{training \& trusted data}
		\label{fig:hp_data}
	\end{minipage}%
	~
	\begin{minipage}[t]{0.45\columnwidth}
		\centering
		\includegraphics[width=.9\columnwidth]{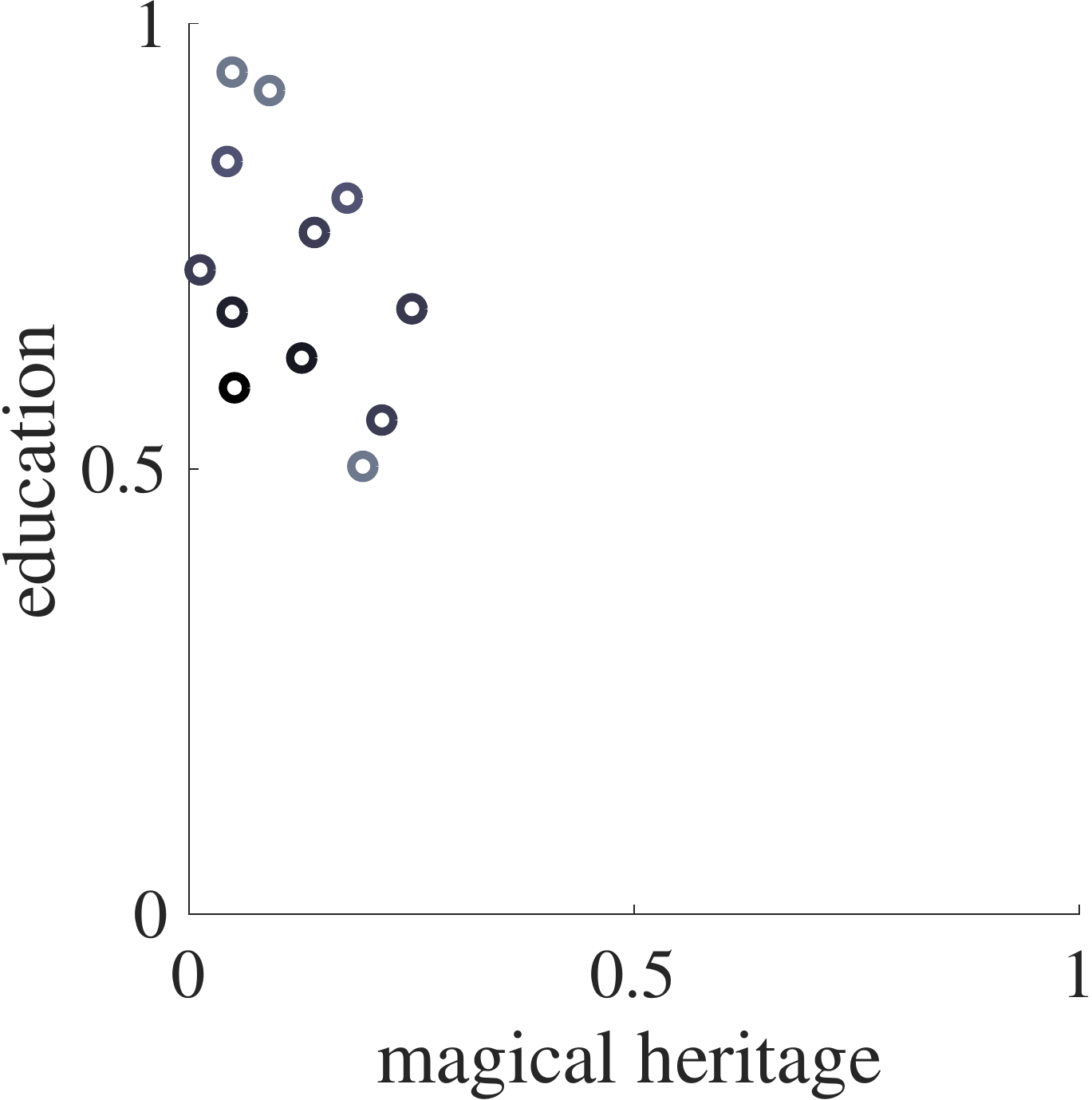}
		\subcaption{\HAPPY flagged bugs}
		\label{fig:hp_duti_flag}
	\end{minipage}
	\caption{Harry Potter Toy Example}
	\label{fig:hp}
\end{figure}

To build intuition, consider a toy example \textit{a la} Harry Potter in Figure~\ref{fig:hp_data}.
Each blue point is a Hogwarts student whose magical heritage ranges from 0 (muggle-born) to 1 (pureblood), and education ranges from 0 (failed school) to 1 (Hermione level).
These blue points form a classification training set, where the class label is `+' (hired by Ministry of Magic after graduation) or `o' (not hired).
This training set shows historical bias against muggle-borns with high education.
Kernel logistic regression trained on the data reflects this bias (black decision boundary).
But suppose we know two more students and how they \emph{should} be classified (the red points) -- the assumption being that a fair decision is based simply on education $\ge 0.5$.
These two points are the trusted items.
Simply training on the trusted items alone will be unsatisfactory -- the boundary will not be flat at education$=0.5$.
Instead, \HAPPY can utilize the trusted items to flag potential bugs in the training set (Fig.~\ref{fig:hp_duti_flag}). 
Darker color represents higher confidence that a training item is a bug.

\section{Training Set Debugging Formulation}
\HAPPY needs three inputs:
\begin{enumerate}
	\item The training set in the form of feature, label pairs $(X,Y)=\{(x_i, y_i)\}_{1:n}$.
The labels $Y$ can be continuous for regression or discrete for classification, and are potentially buggy. 

	\item Trusted items $(\tilde X,\tilde Y)=\{(\tilde x_i, \tilde y_i, c_i)\}_{1:m}$.
These are items verified by domain experts typically at considerable expense. 
The domain expert can optionally specify a confidence $c \ge 0$ for each trusted item.
We assume $m \ll n$ so that the amount of trusted data is insufficient to train a good model.
We do not assume the trusted items are $iid$.
	
	\item The learning algorithm $\A$.  
	In this work, we focus on regularized empirical risk minimizers
	\begin{equation}\label{eq:learner}
	\A(X, Y) = \argmin_{\theta \in \R^p} \frac{1}{n}\sum^{n}_{i=1} \ell(x_i, y_i, \theta) + \lambda \Omega(\theta)
	\end{equation}
	with strongly convex and twice differentiable objective function, such that $\argmin$ returns a unique solution, and the Hessian matrix is always nonsingular.

\end{enumerate}
Conceptually, \HAPPY solves the following optimization problem:
\begin{eqnarray}
&&\argmin_{Y'} \; Distance(Y',Y) \label{eq:concept_obj}\\
&&\mbox{s.t.}\; Predictor = \A(X,Y') \label{eq:learning_constraint} \\
&&\phantom{\mbox{s.t.}}\; Predictor(\tilde X) = \tilde Y \land  Predictor(X) = Y' \label{eq:trust_constraint}
\end{eqnarray}
That is, we find an alternative labeling $Y'$ for the training data, as
close as possible to the original labels $Y$, such that the model
trained with the original feature vectors $X$ and the alternative labels
$Y'$ correctly predicts the labels $\tilde{Y}$ of the trusted
items $\tilde{X}$, and the alternative labeling is self-consistent.
We call the training items for which $y_i'\neq y_i$ the \emph{flagged bugs}; we give them to the domain expert for further inspection to see if they are true bugs.

Our next step is to relax it into a continuous single level optimization problem that can be solved efficiently.

\subsection{Debugging for Regression}
\label{sec:regression}
In regression, $Y'$ and $Y$ are both vectors in $\R^n$.
We define $\delta \equiv Y'-Y$ , and choose $\|\delta\|_1$ as the distance metric to encourage sparsity of $\delta$.
We will denote the predictor $\A(X,Y')$ as $\theta$. Instead of requiring equalities as in~\eqref{eq:trust_constraint}, we relax them by the learner's surrogate loss function, and place them in the objective in the Lagrangian form: 
\begin{eqnarray} 
\min_{\delta \in \R^n,\theta} &&
\frac{1}{m}\sum^{m}_{i=1} c_i \ell(\tilde x_i, \tilde y_i, \theta) \nonumber \\
&& + \frac{1}{n}\sum^{n}_{i=1} \ell(x_i, y_i+\delta_i, \theta )
+ \gamma\frac{\|\delta\|_1}{n} 
\label{eq:Aregression} \\
\mbox{s.t.} &&  \theta = \argmin_{\beta \in \R^p} \frac{1}{n}\sum_{i=1}^n \ell(x_i, y_i+\delta_i,\beta) + \lambda \Omega(\beta).\nonumber  
\end{eqnarray}
where $c_i$'s are the confidence levels assigned to each of the trusted items. 
This is a bilevel optimization problem. 
We now convert the lower level problem to a nonlinear constraint.
Notice that since the lower problem is unconstrained and strongly convex, it can be replaced equivalently with its Karush–Kuhn–Tucker (KKT) condition: 
\begin{equation}
g(\delta, \theta) \equiv \frac{1}{n}\sum_{i=1}^n \nabla_\theta \ell(x_i, y_i + \delta_i, \theta) + \lambda \nabla_\theta \Omega(\theta) = 0.
\label{eq:KKT}
\end{equation}
Now, since $g$ is continuously differentiable and $\frac{\partial g}{\partial \theta}$ is invertible, 
the solution to $g(\delta, \theta) = 0$ defines an implicit function $\theta(\delta)$.
We can now replace $\theta$ with $\theta(\delta)$ in~\eqref{eq:Aregression}, and call the result $\O_\gamma(\delta)$.
Using the fact
$\frac{d \ell}{d \delta} = \frac{\partial \ell}{\partial \delta}+\frac{\partial \ell}{\partial \theta} \frac{\partial \theta}{\partial \delta}$,
we can compute the gradient of $\O_\gamma(\delta)$ as
\begin{eqnarray} 
\nabla_\delta  \O_\gamma &=& \frac{1}{m}\sum^{m}_{i=1} c_i J^\top \nabla_\theta \ell(\tilde x_i, \tilde y_i, \theta) \vert_{\theta(\delta)} \label{eq:gradO} \\
&& + \frac{1}{n}\sum^{n}_{i=1}
\frac{\partial \ell(x_i, y_i+\delta_i, \theta(\delta) )}{\partial \delta_i} \bfe_i  \nonumber \\
&& + \frac{1}{n}\sum^{n}_{i=1}J^\top \nabla_\theta \ell(x_i, y_i + \delta_i, \theta(\delta))
+ \frac{\gamma}{n}\sign(\delta)
\nonumber 
\end{eqnarray} 
where $\bfe_i$ is the $i$th canonical vector and $J$ is defined by the implicit function theorem:
\begin{equation}
J \equiv \frac{\partial \theta}{\partial \delta} = 
-
\begin{bmatrix}
\frac{\partial g_1}{\partial \theta_1} & \cdots & \frac{\partial g_1}{\partial \theta_p} \\
\vdots &  & \vdots \\
\frac{\partial g_p}{\partial \theta_1} & \cdots & \frac{\partial g_p}{\partial \theta_p} 
\end{bmatrix}^{-1}
\begin{bmatrix}
\frac{\partial g_1}{\partial \delta_1} & \cdots & \frac{\partial g_1}{\partial \delta_n} \\
\vdots &  & \vdots \\
\frac{\partial g_p}{\partial \delta_1} & \cdots & \frac{\partial g_p}{\partial \delta_n} 
\end{bmatrix}.
\nonumber
\end{equation}
With $\nabla_\delta  \O_\gamma$ we then solve~\eqref{eq:Aregression} with a gradient method.

\subsection{Debugging for Classification}
Let there be $k$ classes.
To avoid a combinatorial problem, we relax the optimization variable to the $k$-probability simplex $\Delta_k$.
Concretely, we first augment the learner $\A$ so that it takes \emph{weighted} training items:
\begin{equation}
\theta = \argmin_\beta \frac{1}{n}\sum_{i=1}^n \sum_{j=1}^k w_{ij} \ell(x_i, j,\beta) + \lambda \Omega(\beta)\label{eq:MNLog}
\end{equation}
where $w_i \in \Delta_k$, that is, $w_{ij}\ge 0$ and $\sum_{j=1}^k w_{ij}=1$, $\forall i, j$.
The original learner~\eqref{eq:learner} can be recovered with $w_i = \bfe_{y_i}, \forall i$.

We then represent the $i$th proposed new class label $y'_i$ by $\delta_i \in \Delta_k$. Note that $\delta_i$ here represents the \emph{new} set of labels, not a difference in labels.
One way to measure the distance to the old label $y_i$ is $1-\delta_{i,y_i}$, namely the probability mass siphoned away from the old label.
We thus obtain a bilevel optimization problem with continuous variables similar to~\eqref{eq:Aregression}:
\begin{eqnarray}
\label{eq:Aclassification}
\min_{\delta\in\Delta_k^n, \theta} && \frac{1}{m}\sum^{m}_{i=1} c_i \ell(\tilde x_i, \tilde y_i, \theta)\\
&&+\frac{1}{n}\sum^{n}_{i=1} \sum_{j=1}^k \delta_{ij} \ell(x_i, j, \theta )
+\frac{\gamma}{n}\sum_{i=1}^n (1-\delta_{i, y_i})\nonumber\\
\mbox{s.t.} && \theta = \argmin_\beta \frac{1}{n}\sum_{i=1}^n \sum_{j=1}^k \delta_{ij} \ell(x_i, j,\beta) + \lambda \Omega(\beta). \nonumber
\end{eqnarray}
Finally, we go through similar steps as in regression:
replacing the lower problem with its KKT condition;
defining an implicit function $\theta(\delta)$;
obtaining the objective $\O_\gamma(\delta)$ of~\eqref{eq:Aclassification};
and computing its gradient $\nabla_\delta  \O_\gamma$ using implicit function theorem.

In the classification case, the KKT condition can be written as
\begin{equation}\label{eq:KKTclassification}
g(\delta, \theta) \equiv \frac{1}{n}\sum_{i=1}^n \sum_{j=1}^k \delta_{ij} \nabla_\theta\ell(x_i, j,\beta) + \lambda \nabla_\theta \Omega(\theta) = 0.
\end{equation}
We then linearize the soft label as $\delta = (\delta_{11},\cdots,\delta_{1k},\cdots,\delta_{nk})$.
By assumptions, $g$ is continous differentiable and $\frac{\partial g}{\partial \theta}$ is invertible, and therefore
the solution to $g(\delta, \theta) = 0$ defines an implicit function $\theta(\delta)$ and similarly
\begin{equation}
J = \frac{\partial \theta}{\partial \delta} = 
-
\begin{bmatrix}
\frac{\partial g_1}{\partial \theta_1} & \cdots & \frac{\partial g_1}{\partial \theta_p} \\
\vdots &  & \vdots \\
\frac{\partial g_p}{\partial \theta_1} & \cdots & \frac{\partial g_p}{\partial \theta_p} 
\end{bmatrix}^{-1}
\begin{bmatrix}
\frac{\partial g_1}{\partial \delta_{11}} & \cdots  \frac{\partial g_1}{\partial \delta_{1k}} 
&\cdots  \frac{\partial g_1}{\partial \delta_{nk}}\\
\vdots &   \vdots & \vdots\\
\frac{\partial g_p}{\partial \delta_{11}} & \cdots  \frac{\partial g_p}{\partial \delta_{1k}} 
&\cdots  \frac{\partial g_1}{\partial \delta_{nk}}
\end{bmatrix}.
\end{equation}

Then the optimization problem becomes a nonlinear optimization problem with bound inequality constraints and linear equality constraints.
\begin{align} 
\argmin_{\delta\in[0,1]^k} \,&
\frac{1}{m}\sum^{m}_{i=1} c_i \ell(\tilde x_i, \tilde y_i, \theta(\delta))
+\\
&\frac{1}{n}\sum^{n}_{i=1} \sum_{j=1}^k \delta_{ij} \ell(x_i, j, \theta(\delta) )
+\gamma \frac{1}{n}\sum_{i=1}^n (1-\delta_{i, y_i})\nonumber\\
& \mbox{s.t. }\sum_{j=1}^k \delta_{ij} = 1, \forall i = 1,\cdots, n
\end{align} 
Similarly denote the objective as $\O_\gamma(\delta)$.
We may write down the gradient as
\begin{eqnarray} 
\nabla_\delta  \O_\gamma &=& 
\frac{1}{m}\sum^{m}_{i=1} c_i J^\top \nabla_\theta \ell(\tilde x_i, \tilde y_i, \theta) \vert_{\theta(\delta)} \\
&&+\frac{1}{n}\sum^{n}_{i=1} \sum_{j=1}^k \ell(x_i, j, \theta(\delta) ) \bfe_{ij} \nonumber\\
&&+\frac{1}{n}\sum^{n}_{i=1} \sum_{j=1}^k\delta_{ij} J^\top\nabla_\theta \ell(x_i, y_i, \theta) \vert_{\theta(\delta)}-
\gamma \frac{1}{n}\sum_{i=1}^n \bfe_{iy_i}\nonumber
\end{eqnarray} 

\subsection{The \HAPPY Algorithm}
\LinesNumbered
\begin{algorithm}[ht]
	\caption{\HAPPY\label{alg}}
	\SetKwInOut{Input}{Input}
	\SetKwInOut{Output}{Output}
	\Input{Training set $(X,Y)$, trusted items $(\tilde X, \tilde Y, c)$, \newline learner $\A$, examination budget $b\leq n$;}

	Initialize $t = 0$, $\delta^{(0)} = 0$ in regression or $Y$ in classification, $n_{flag} = 0$\;
	Initialize  $\gamma^{(0)} = \max_i |\nabla_\delta  \O_{\gamma = 0}(\delta^{(0)})_i|$ in regression, or $\gamma^{(0)} = \max_i \nabla_\delta  \O_{\gamma = 0}(\delta^{(0)})_{i,y_i}$ in classification\;
	\While { $n_{flag} \leq b$}{
		$t = t+1$\;
		$\gamma^{(t)} = \gamma^{(t-1)} /2$\;
		$\delta^{(t)} = \argmin\O_{\gamma^{(t)} }(\delta)$, initialized at $\delta^{(t-1)}$\;

		$F^{(t)}  = \{i\mid \delta^{(t)}_{i}\neq 0\}$ in regression, or\\
		$F^{(t)}  = \{i\mid \argmax_j(\delta^{(t)}_{ij})\neq y_i\}$ in classification\;

		$n_{flag} = |\cup_{s=1}^t F^{(s)} |$\;
	}
	\Output{$(\gamma_1, \delta^{(1)}), \cdots, (\gamma_t,\delta^{(t)})$;
	}
\end{algorithm}

We now present a unified \HAPPY algorithm for debugging both regression and classification, see Algorithm~\ref{alg}.
As part of the input, the domain expert can specify an examination budget $b$, the preferred total number of flagged bugs that they are willing to check.
Recall that the debugging formulations~\eqref{eq:Aregression} and~\eqref{eq:Aclassification} have a sparsity parameter $\gamma$.
\HAPPY automatically chooses a sequence of $\gamma$'s, runs the corresponding debugging formulation, and flags potential bugs $F^{(t)} $ by line 7 or 8, respectively.
$F^{(1)} , F^{(2)} , \ldots$ are not necessarily nested subsets.
\HAPPY accumulates all previously flagged bugs by $\cup_{s=1}^t F^{(s)} $, and only stops if its size exceeds the examination budget $b$.

\HAPPY outputs the sequence of sparsity parameters $\gamma$ and solutions $\delta$, from which the $F$'s can be recovered.
This is helpful for the domain expert to investigate the flagged bugs.
Specifically, \HAPPY's output induces a \emph{ranking} over all flagged bugs.
Bugs are ranked by the earliest time they appear in the sequence $F_1, F_2, \ldots$. 
When two bugs first appear at the same time, the tie is broken by comparing the deviation of $\delta$ from the initial value. Larger deviation = earlier order. 
Furthermore, the value $\delta$ suggests the fix, namely what regression output value or class label \HAPPY thinks the flagged bug should have had.

\HAPPY chooses the $\gamma$ sequence as follows.
It starts with the largest $\gamma$ that returns a nontrivial solution $\delta$, namely $\delta \neq \mathbf{0}$ in regression and $\delta\neq Y$ in classification. 
One can show that these are sufficient conditions for a nontrivial $\delta$ solution:
In the regression setting, $\gamma^{(0)}  = \max_i | \left[ \nabla_\delta  \O_{\gamma = 0}(\mathbf 0) \right]_i|$, where $\nabla_\delta  \O_{\gamma = 0}$ is the gradient of the debugging objective with $\gamma = 0$, taken at $\delta = \mathbf{0}$.
In the multiclass classification setting, $\gamma^{(0)}  = \max_i \left[ \nabla_\delta  \O_{\gamma = 0}(Y) \right]_{i,y_i}$, 
where $\nabla_\delta  \O_{\gamma = 0}$ is similarly the gradient of the debugging objective with $\gamma = 0$, taken at the initial value $\delta = Y$ . 
The $i,y_i$-th entry of the gradient is the probability mass assigned to the original label $y_i$.  A positive entry will result in optimization taking a gradient step, thus returning a nontrivial solution.

\HAPPY utilizes continuation method to speed up optimization. Specifically,  each iteration $t$ is initialized using the previous solution $\delta^{(t-1)}$.
Moreover, \HAPPY uses a linear approximation 
$\theta^{(t-1)} + \frac{\partial \theta}{\partial\delta}
(\delta^{(t)}-\delta^{(t-1)})
\vert_{\delta^{(t-1)}}
$ to initialize the computation of $\theta(\delta^{(t)})$.

\section{Experiments}
To the best of our knowledge, no machine learning debugging repository is publicly available. 
That is, no data set has provenance like ``item $i$ had the wrong label $y$; it should have label $z$'' for research in debugging.
Curating such a repository will be useful for trustworthy machine learning research.
Software debugging repository in the programming language community, such as the Siemens test suite~\cite{hutchins1994experiments} and BugBench~\cite{lu2005bugbench}, are good examples to follow.
In this paper, we had to simulate -- as plausible as possible -- true bugs on the real data sets.

\subsection{Learner $\A$ Instantiation in \HAPPY}
\HAPPY is a \emph{family} of debuggers depending on the learner $\A$~\eqref{eq:learner}.
For illustration, in our experiments we let $\A$ be kernel ridge regression for regression and multiclass kernel logistic regression for classification, both with RBF kernels.
We recall the relevant learner loss $\ell$ and regularizer $\Omega$ below; the derivation of the corresponding debugging objective $\O_\gamma$ and gradient $\nabla_\delta \O_\gamma$ is straightforward but tedious, and is left for a longer version.
Extension to other learners can be done similarly.

\subsubsection{Kernel Ridge Regression}
Let $(X,Y)=\{(x_i, y_i)\}_{1:n}$ be the training data, and $K = [k(x_i,x_j)]_{n\times n}$ be the kernel matrix. Denote by $K_i$ the $i$-th column of $K$, and let $\alpha\in\R^n$ be the learning parameter. Kernel ridge regression can be written in the form of regularized ERM~\eqref{eq:learner}, where $\ell(x_i,y_i,\alpha) = (y_i - K_i\alpha)^2$ and $\Omega(\alpha) = \frac{1}{2}\alpha^\top K\alpha$.

In the case of Kernel Ridge Regression, the learning problem in fact yield closed-form solution. Therefore, we can spare the steps of computing the KKT condition and applying influence function, and directly plug the learning solution into the objective. In particular, given that
\begin{equation}
\alpha = \argmin_{\beta \in \R^n} \frac{1}{n}\sum_{i=1}^n (y_i + \delta_i - K_i\beta)^2 + \lambda\beta^\top K\beta
\end{equation}
The objective $\O_\gamma$ can be written as
\begin{align}
\min_{\delta \in \R^n} &
\frac{1}{m}\left[C\left(\tilde y - \tilde K(K+n\lambda I_n)^{-1}(y+\delta)\right)\right]^2+\\
&\frac{1}{n}\left[(y + \delta) - K(K+n\lambda I_n)^{-1}(y+\delta)\right]^2+\\
&\gamma\frac{\|\delta\|_1}{n},
\label{eq:KRR1}
\end{align}
where $C = \diag(\sqrt{c_1},\cdots,\sqrt{c_m})$. If we denote $A = \tilde K(K+n\lambda I_n)^{-1}$, $B =  K(K+n\lambda I_n)^{-1} - I_n$ and $C' = \begin{bmatrix}
\frac{1}{\sqrt{m}}C & 0\\
0 & \frac{1}{\sqrt{n}}I_n
\end{bmatrix}$, then equation~\eqref{eq:KRR1} can be simplified to
\begin{equation}
\min_{\delta \in \R^n} 
\left\|C'\left(
\begin{bmatrix}
A\\B
\end{bmatrix}
\delta-
\begin{bmatrix}
\tilde y - Ay\\-By
\end{bmatrix}
\right)\right\|^2
+ \gamma\frac{\|\delta\|_1}{n},
\label{eq:KRR1}
\end{equation}
i.e. a weighted lasso problem that is in fact convex and yields unique solution.
\subsubsection{Multiclass Kernel Logistic Regression with Ridge Regularization}
Let $\alpha\in\R_{n\times k}$ be the learning parameter, and denote by $\alpha_j$ the $j$-th column of $\alpha$. 
Kernel logistic regression can be written in the form of weighted learner~\eqref{eq:MNLog}:
\begin{align}
\theta = &\argmin_\alpha -\frac{1}{n}\sum_{i=1}^n\sum_{j=1}^k \delta_{ij}K_i^\top\alpha_j \nonumber \\
& + \frac{1}{n}\sum_{i=1}^n\log\left(\sum_{j=1}^k \exp(K^\top_i \alpha_j)\right) + \frac{\lambda}{2}\sum_{j=1}^k\alpha^\top_jK\alpha_j \nonumber
\end{align}
where 
$\ell(x_i,j,\alpha) = - K_i^\top\alpha_j + \log\left(\sum_{j=1}^k \exp(K_i^\top \alpha_j)\right)$ and $\Omega(\alpha) = \frac{1}{2}\sum_{j=1}^k\alpha^\top_jK\alpha_j$.

The debugging objective $\O_\gamma$ can be instantiated as
\begin{align} 
\argmin_{\delta\in[0,1]^k} &
\frac{1}{m}\sum^{m}_{i=1} c_i \left[- \tilde K_i^\top\alpha_{\tilde y_i} + \log\left(\sum_{j=1}^k \exp(\tilde K_i^\top \alpha_j)\right)\right]+\\
&\frac{1}{n}\sum^{n}_{i=1}  \left[- \sum_{j=1}^k \delta_{ij}K_i^\top\alpha_j + \log\left(\sum_{j=1}^k \exp(K_i^\top \alpha_j)\right)\right]+\\
&\frac{\gamma}{n}\sum_{i=1}^n (1-\delta_{i, y_i})\\
\mbox{s.t. }&\sum_{j=1}^k \delta_{ij} = 1, \forall i = 1,\cdots, n
\end{align}
Now we need to compute the gradient $\nabla_\delta  \O_\gamma$. To do that, we first need to compute the gradient and Hessian of the loss function. Denote $\bfe_j$ the linearization of a $n\times k$ matrix with all entries zero except the $j$-th row, $K_i'$ the linearization of a $n\times k$ matrix with each column equal to $K_i$, and $*$ the entry-wise multiplication. Then, one can compute the gradient and Hessian as follow:
\begin{equation}
\nabla_\alpha \ell(x_i,y_i,\alpha) = K_i' *\bfe_{y_i}
+ \frac{\sum_{j=1}^k \left(\exp(K_i^\top \alpha_j)K_i'*\bfe_j\right)}{\sum_{j=1}^k \exp(K_i^\top \alpha_j)}
\end{equation}
\begin{align}
&H_\alpha \ell(x_i,y_i,\alpha) =
\frac{\sum_{j=1}^k \left(\exp(K_i^\top \alpha_j)(K_i'*\bfe_j)(K_i'*\bfe_j)^\top\right)}{\sum_{j=1}^k \exp(K_i^\top \alpha_j)}-\\	
&\frac{\left[\sum_{j=1}^k \left(\exp(K_i^\top \alpha_j)K_i'*\bfe_j\right)\right]\left[\sum_{j=1}^k \left(\exp(K_i^\top \alpha_j)K_i'*\bfe_j\right)\right]^\top}{\left(\sum_{j=1}^k \exp(K_i^\top \alpha_j)\right)^2}
\end{align}
Given these, and denote $K'$ as the $nk\times nk$ matrix with $k$ K's along the diagonal, we can then compute the KKT condition as
\begin{align}
g(\delta,\alpha)= &\frac{1}{n}\sum_{i=1}^n \left[\left(\sum_{j=1}^k\delta_{ij}K_i'*\bfe_j\right)
+ \frac{\sum_{j=1}^k \left(\exp(K_i^\top \alpha_j)K_i'*\bfe_j\right)}{\sum_{j=1}^k \exp(K_i^\top \alpha_j)}
\right]\\
&+\lambda K'\alpha
\end{align}
and
\begin{equation}
J = \frac{\partial \alpha}{\partial \delta} = 
-
\begin{bmatrix}
\vert &   \vert & \vert\\
\frac{\partial g}{\partial \alpha_{11}} & \cdots  \frac{\partial g}{\partial \alpha_{1k}} 
&\cdots  \frac{\partial g}{\partial \alpha_{nk}}\\
\vert &   \vert & \vert\\
\end{bmatrix}^{-1}\cdot
\frac{\partial g}{\partial \alpha}
\end{equation}
where $\frac{\partial g}{\partial \delta_{ij}} = \frac{1}{n}K'_i*\bfe_j$. Observe that $H_\alpha \ell(x_i,y_i,\alpha)$ is independent of $y_i$, so we can simplify $\frac{\partial g}{\partial \alpha}$ as
\begin{eqnarray}
\frac{\partial g}{\partial \alpha} = \frac{1}{n}\sum_{i=1}^n H_\alpha\ell(x_i,y_i,\alpha) + \lambda K'
\end{eqnarray}

In all our experiments, the learner's hyperparameters are set by 10-fold cross validation on the original training data, and confidence levels on all trusted items $c$ are set to 100.

\subsection{Baseline Debugging Methods for Comparison}
We compare \HAPPY with three baselines: influence function (INF)~\cite{koh2017understanding}, 
Nearest Neighbor (NN), and
Label Noise Detection (LND)~\cite{bhadra2015}.

\subsubsection{Influence Function}\cite{koh2017understanding} describes how perturbing a training point changes the prediction on a tested point. 
The influence of training labels on the trusted items can be written as
\begin{eqnarray}
I =  \frac{1}{m}\sum^{m}_{i=1}J^\top \nabla_\theta \ell(\tilde x_i, \tilde y_i, \theta) \vert_{\theta(\delta)}
\end{eqnarray}
This is in fact the first term of our objective gradient $\nabla_\delta  \O_\gamma$ (assuming $c_i=1, \forall i$).
Therefore, one can view influence function as a first-order approximation to a simplified version of \HAPPY. 
Intuitively, a positive influence implies that decreasing the $y$ value will reduce the loss of trusted items,
 while a negative influence implies that one should increase the corresponding $y$ value to achieve the same result. In regression, INF prioritizes training points with larger absolute values of influence. 
In classification, each possible label of a training point will have an influence. 
INF will flag training points with positive influence on their original label, and prioritize ones with larger influence value. In suggesting fixes for each flagged training point, INF will suggest the label with the largest negative influence value.

Since INF is a first-order approximation to \HAPPY, we expect INF to be inferior for debugging due to nonlinearity between labels and predictions.  This is confirmed by our experiments.

\subsubsection{Nearest Neighbor}
NN is a simple heuristic: 
In regression, NN flags training points based on the Euclidean distance (after normalizing each feature dimension separately) to the closest trusted item. 
In classification, NN flags each training point whose label differs from the label of the closest trusted point, and prioritizes them by this distance. 
When asked for a suggested label fix, it recommends the label of closest trusted item.

\subsubsection{Label Noise Detection (Oracle)}\cite{bhadra2015} uses a Gaussian kernel density estimator as the consistency metric to define a combinatorial optimization problem:
\begin{eqnarray}
\argmax_{\eta\in\{1,-1\}^n} \eta^\top Q\eta.
\end{eqnarray}
where $Q_{ij} = y_iy_jK_h(x_i,x_j)$, $\eta_i = 1$ represents a correct original label, and $\eta_i = -1$ a bug.
LND is only for binary labels.
 The algorithm finds the best relabeling $\eta$ that maximize this mutual consistency metric. 
In the case that there are expert verified data, e.g. our trusted items, LND can incorporate them as constraints to reduce the search space. 
However, LND requires the user to provide the number of positive mislabeled items and the number of negative mislabeled items.
In practice such information is usually unavailable, but in our experiments we provide LND with the ground truth numbers as a best-case scenario.
For this reason we label this baseline LND (Oracle).
Also note that LND applies only to binary classification problems, so we omit LND from baselines for our regression and multiclass classification experiments.

\subsection{Toy Data: Harry Potter (Cont.)}
\begin{figure}[t!]
	\centering
	\begin{minipage}[t]{0.45\columnwidth}
		\centering
		\includegraphics[width=.9\columnwidth]{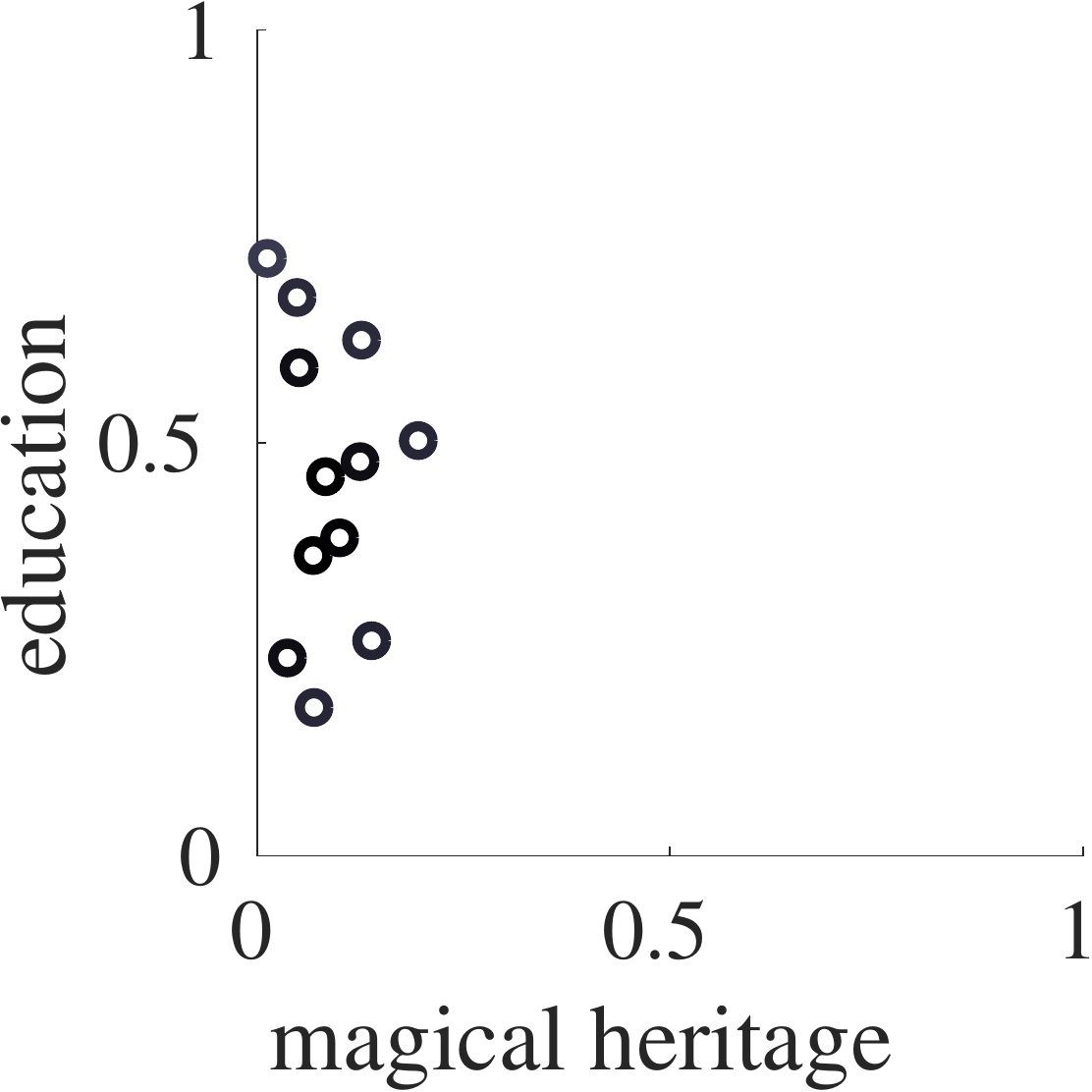} 
		\subcaption{INF flagged bugs}
		\label{fig:hp_inf_flag}
	\end{minipage}%
	~
	\begin{minipage}[t]{0.45\columnwidth}
		\centering
		\includegraphics[width=.9\columnwidth]{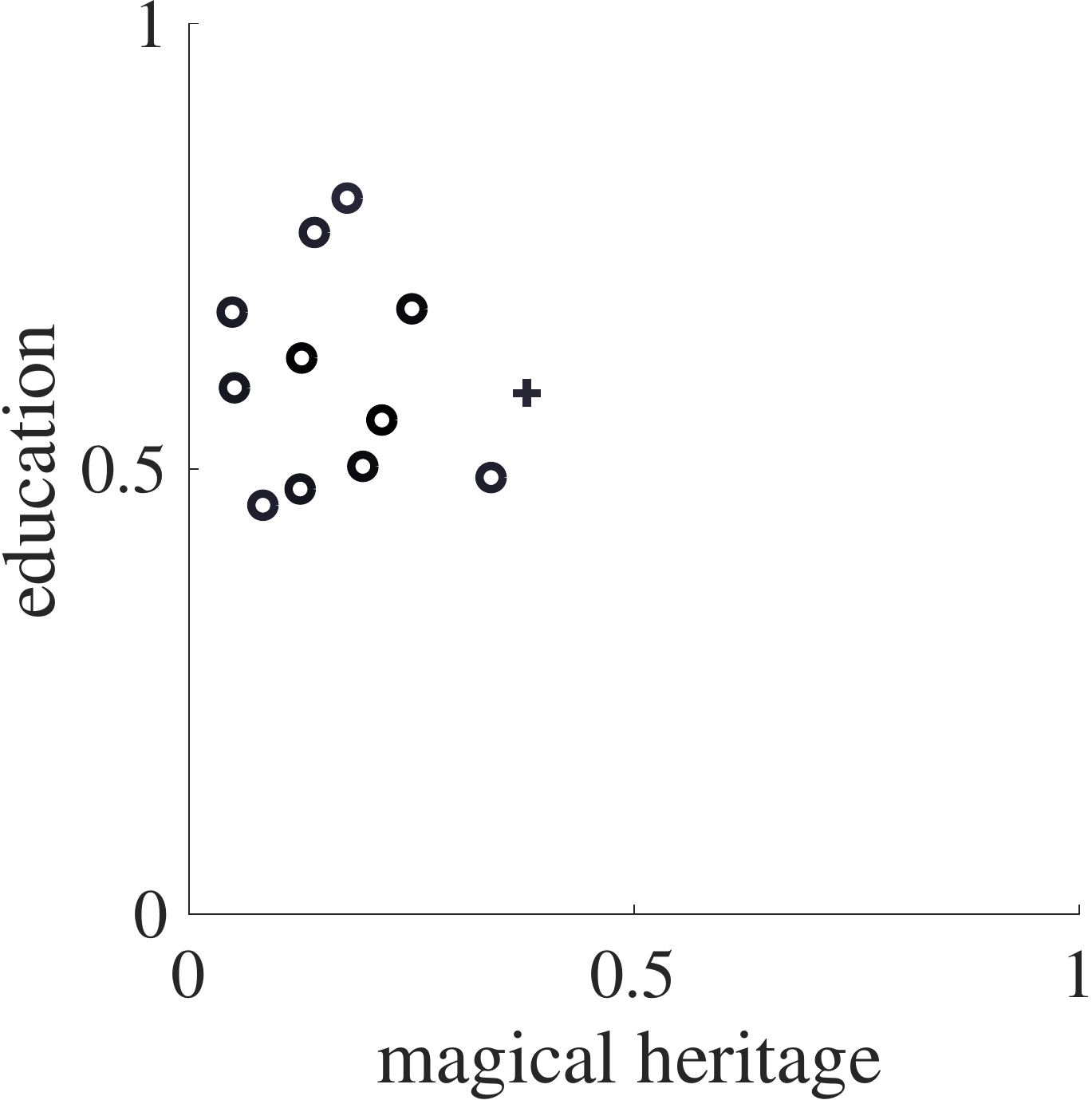}
		\subcaption{NN flagged bugs}
		\label{fig:hp_nn_flag}
	\end{minipage}
	
	\begin{minipage}[t]{0.45\columnwidth}
		\centering
		\includegraphics[width=.9\columnwidth]{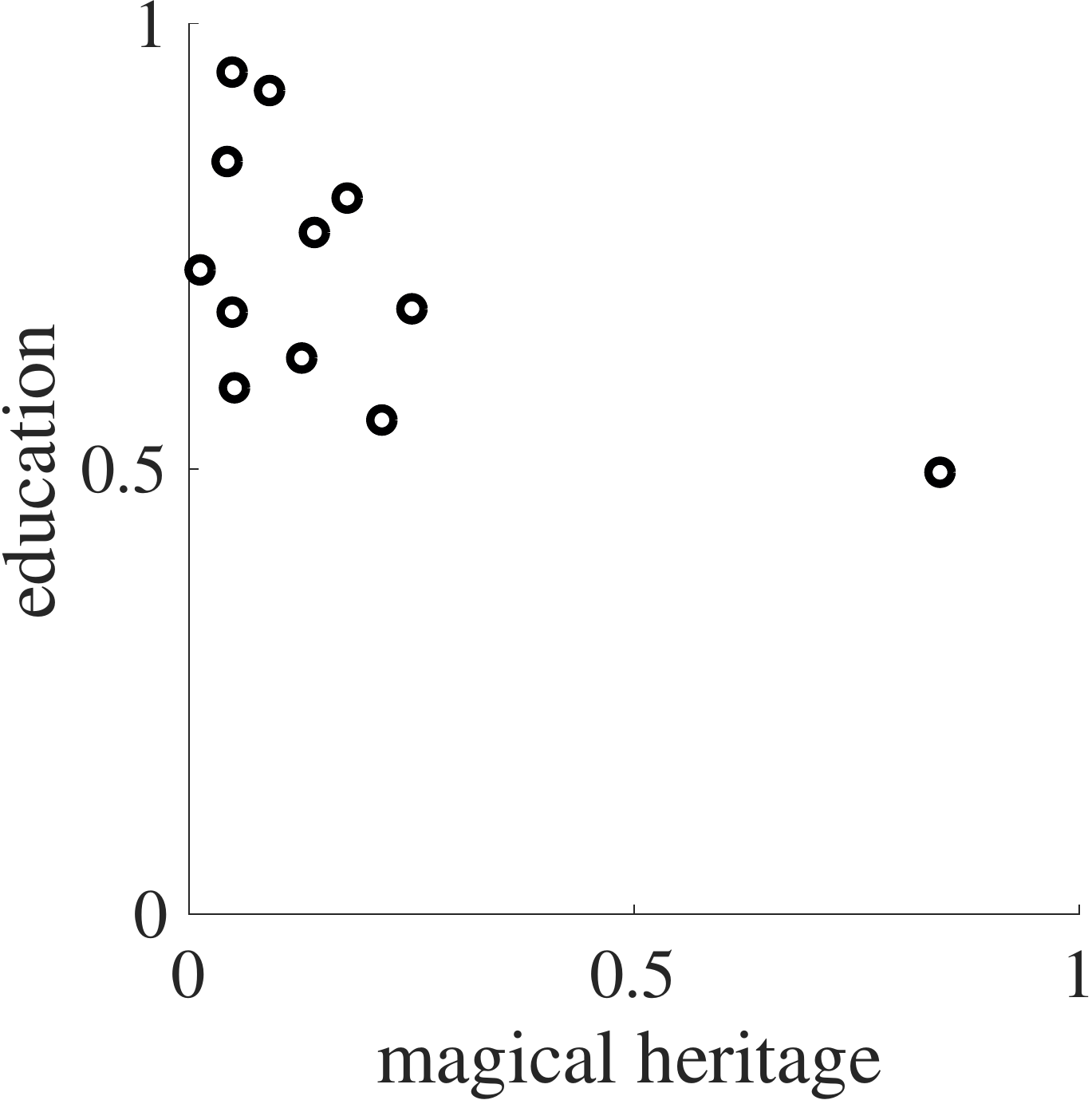} 
		\subcaption{LND flagged bugs}
		\label{fig:hp_lnd_flag}
	\end{minipage}%
	~
	\begin{minipage}[t]{0.45\columnwidth}
		\centering
		\includegraphics[width=.9\columnwidth]{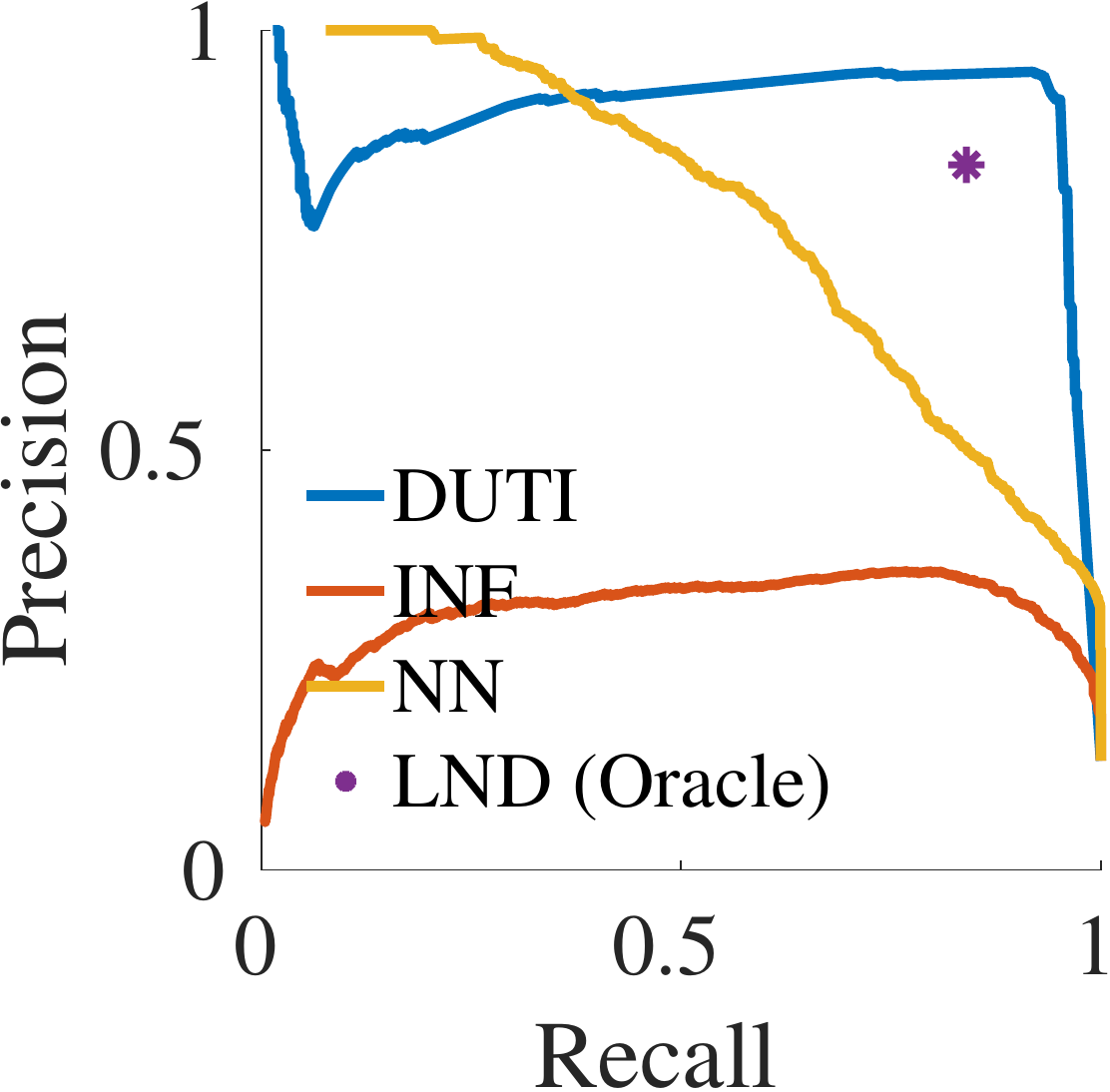}
		\subcaption{Average PR Curves}
		\label{fig:hp_pr}
	\end{minipage}
	\caption{Harry Potter Toy Example, Continued}
	\label{fig:hp2}
\end{figure}
We can now fully define the Harry Potter toy example.
The true boundary is the line education$=0.5$.
A training point whose label disagrees with the true boundary is considered a bug.
Thus the true bugs are all the ``not hired'' training points on the upper left.
Figure~\ref{fig:hp_duti_flag} shows that \HAPPY is able to flag  true bugs with a budget $b=12$. 
Figure~\ref{fig:hp2}(a--c) show the top 12 training items flagged by INF, NN and LND, respectively.
INF and NN flagged some training points below the true boundary, which they should not. 
LND, thanks to the oracle information, is able to flag mostly true bugs but also produced one false positive. 
By varying how many training points we ask each method (other than LND) to flag, we can produce a precision-recall (PR) curve with respect to true bugs for that method.
LND is represented by a single point on the PR curve, because it flags only a fixed number of points due to the oracle information.
Figure~\ref{fig:hp_pr} shows the average PR curves from 100 random runs, in which the training data are randomly drawn with trusted items fixed.
Overall, \HAPPY dominates the baseline methods.

\subsection{Real Data: German Loan Application}
We study the UCI German Loan data set, which has been used in recent work on algorithmic fairness~\cite{zemel2013learning,feldman2015certifying}.
Prior work suggested a systematic bias of declining applicants younger than 25. 
We now detail the way we simulate true bugs on German Loan.
Throughout the learning and debugging process we remove the age attribute.

\begin{figure*}[t!]
	\centering
	\begin{minipage}[t]{0.45\columnwidth}
		\centering
		\includegraphics[width=.9\columnwidth]{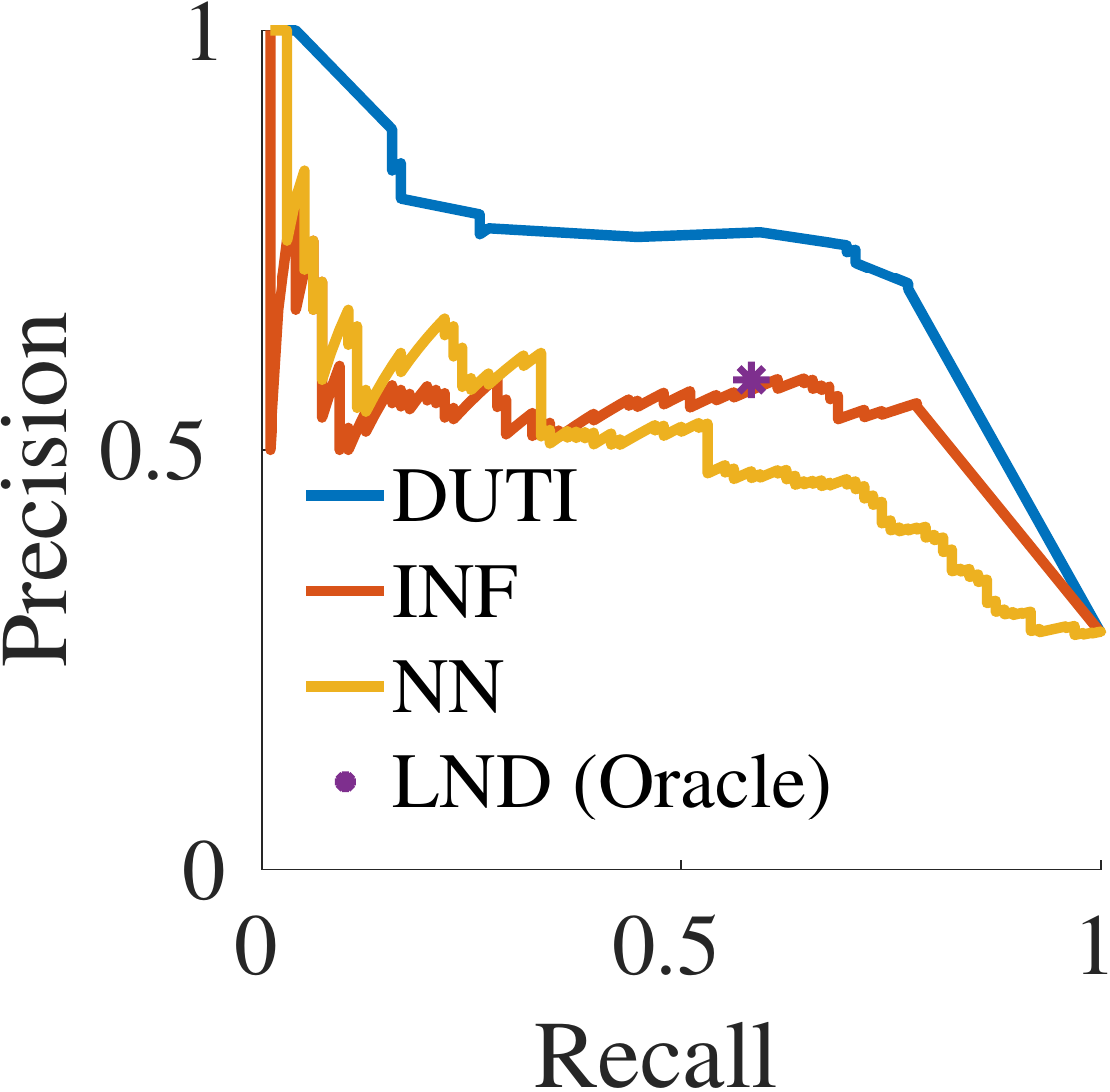} 
		\subcaption{German Loan PR Curves}
		\label{fig:credit_pr}
	\end{minipage}%
	~
	\begin{minipage}[t]{0.45\columnwidth}
		\centering
		\includegraphics[width=.9\columnwidth]{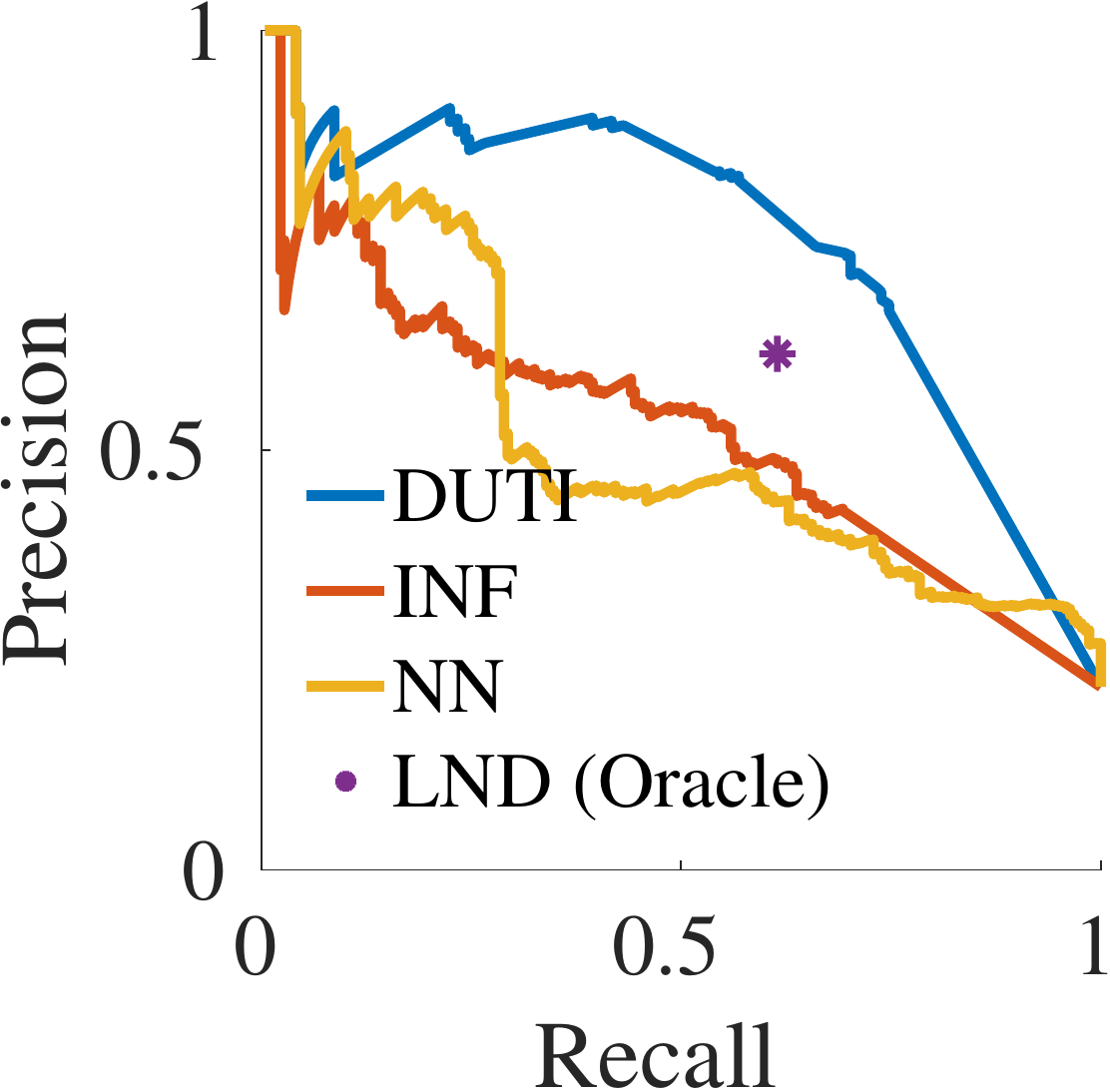}
		\subcaption{Adult Income PR Curves}
		\label{fig:adult_pr}
	\end{minipage}
	\begin{minipage}[t]{0.45\columnwidth}
		\centering
		\includegraphics[width=.9\columnwidth]{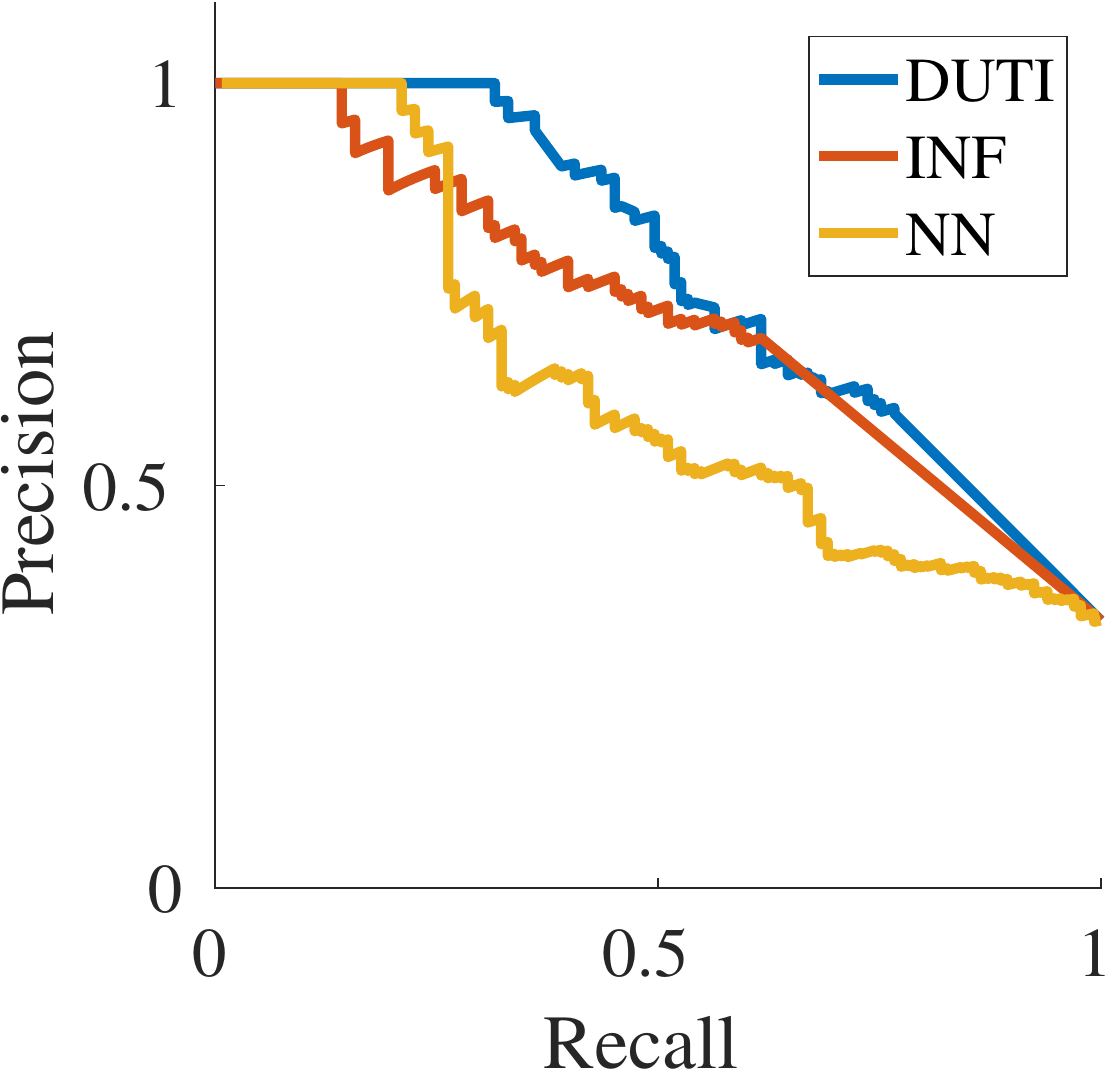} 
		\subcaption{Digits PR Curves}
		\label{fig:digit_pr}
	\end{minipage}%
	~
	\begin{minipage}[t]{0.45\columnwidth}
		\centering
		\includegraphics[width=.9\columnwidth]{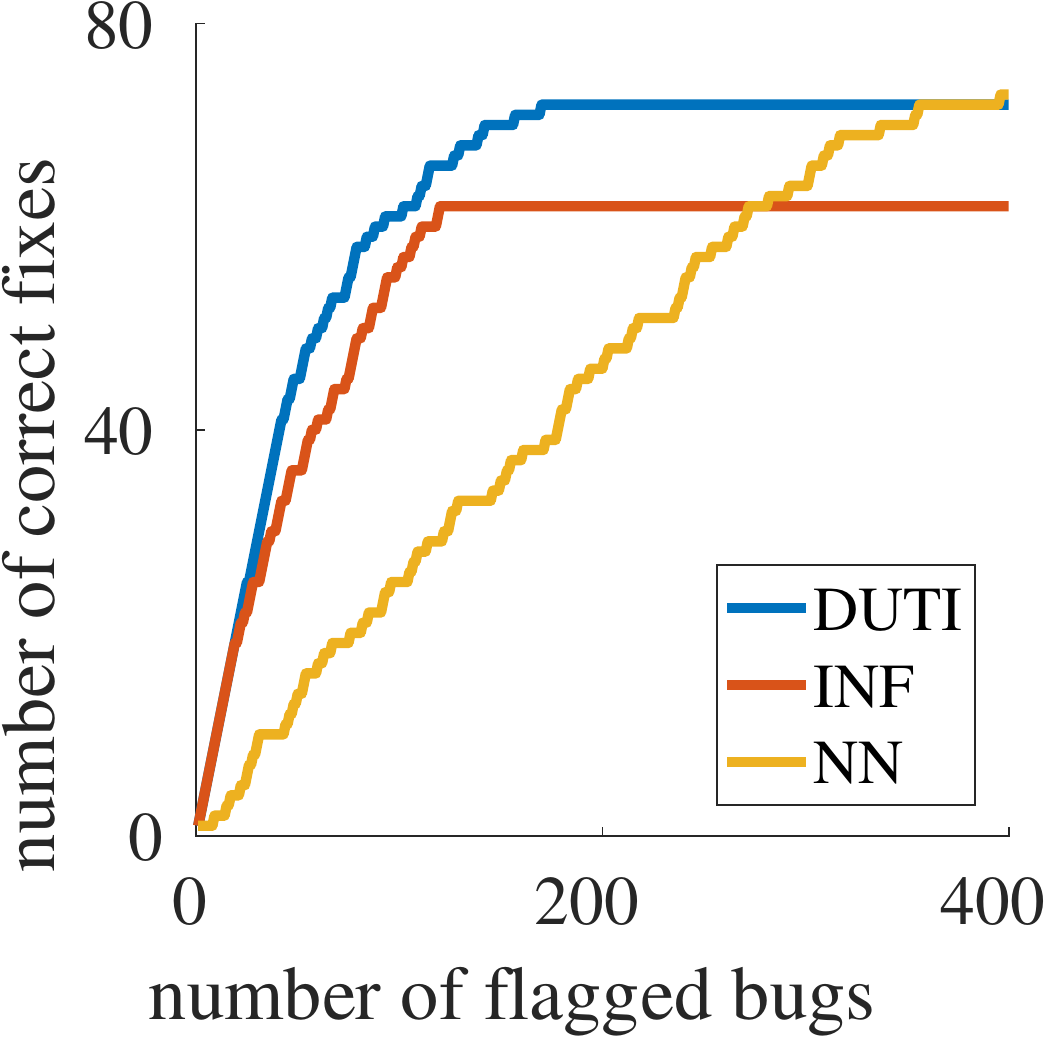}
		\subcaption{Digits suggested fixes}
		\label{fig:digit_fix_pr}
	\end{minipage}
	\caption{Real Data Experiments}
	\label{fig:real}
\end{figure*}

Step 1. The original data set consists of 1000 applicants with 190 young (age$\leq 25$) and 810 old (age$>25$). We partition the dataset into three subsets A,B,C.  A contains 20 random young and 20 random old applicants. B contains the remaining 170 young and another 170 random old applicants. C contains the remaining 620 old applicants.

Step 2.
We use group C to train a loan decision classifier $f^*$, and use it as the ground truth concept.

Step 3.
We relabel the applicants in group A using $f^*$, and treat the relabeled group A as trusted items.

Step 4.
Group B with the original UCI label is treated as the buggy training set. 
Whenever $Y$ disagrees with $f^*(X)$, that training point is considered a bug. This methodology results in 96 bugs. 

Figure~\ref{fig:credit_pr} compares the PR curves of the four debugging methods.
\HAPPY clearly dominates other methods.

\subsection{Real Data: Adult Income}
Another dataset often used in algorithmic fairness is UCI Adult Income~\cite{kohavi1996scaling,kamishima2011fairness}. The task is to classify whether an individual receives an annual income of $\geq$ 50k. Prior work suggests that the data set contains systematic biases against females.
In this experiment, we simulate true bugs based on such biases. Throughout the learning and debugging process we remove the gender attribute and `wife \& husband' values in the family relationship attribute.

Similar to the German Loan Dataset, we randomly subsample 3 disjoint subsets A,B,C. A contains 20 random male and 20 random female applicants. B contains 500 random males and 500 random female applicants. C contains 2000 random male applicants.
We apply exactly the same steps 2,3,4 as in the German Loan data set. This process results in 218 bugs.

Figure~\ref{fig:adult_pr} compares the PR curves of the four debugging methods and again \HAPPY dominates other methods.

\subsection{Real Data: Handwritten Digits}

\begin{table*}[t]
\begin{center}
\begin{tabular}{ |c|c|c|c|c|c|c|c|c|c|c| } 
 \hline
$X$
       &
\includegraphics[width=0.058\textwidth]{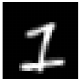}
       &
\includegraphics[width=0.058\textwidth]{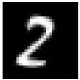}
       &
\includegraphics[width=0.058\textwidth]{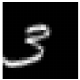}
       &
\includegraphics[width=0.058\textwidth]{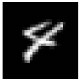}
       &
\includegraphics[width=0.058\textwidth]{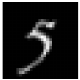}
       &
\includegraphics[width=0.058\textwidth]{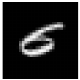}
       &
\includegraphics[width=0.058\textwidth]{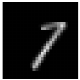}
       &
\includegraphics[width=0.058\textwidth]{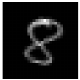}
       &
\includegraphics[width=0.058\textwidth]{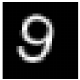}
       &
\includegraphics[width=0.058\textwidth]{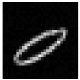}
\\ 
  \hline
$Y = f^*(\mathrm{blur}(X))$ & 3 & 7 &1&8&3&4&1&3&8&8 \\ 
  \hline
  \hline
DUTI&1&2&3&7&5&6&7&8&--&0 \\ 
 \hline
INF&--&--&5&--&5&--&7&0&--&0\\ 
 \hline
NN&1&--&5&7&8&6&4&0&--&5\\ 
 \hline
\end{tabular}
\end{center}
\caption{Selected images with buggy training labels, and the debugging actions on them}
\label{tab:digits}
\end{table*}

In this section, we evaluate the debugging methods on a 10-class handwritten digit recognition problem~\cite{MatlabDeepDigit}.
The digit images are generated by applying random affine transformations to digit images created using different fonts.
In multiclass classification, flagging the true bugs and suggesting the correct fixes are no longer equivalent, so we present evaluations on both criteria.
Unlike the previous two experiments, we now simulate a contamination mechanism to generate buggy labels, while the original labels are considered clean.

Step 1. We randomly set aside 5000 data points, 500 per digit, to train a deepnet consisting of an autoencoder layer followed by a softmax layer,
achieving a cross validation accuracy of 98\%. Denote the trained neural net by $f^*$.

Step 2. Among the rest, we randomly sample 400 data points, 40 per digit, to be the training set $X$. We then blur the images heavily with a Gaussian filter, 
and classify on the blurred images with the trained deepnet. 
The intention is to simulate the generation of buggy labels by a human annotator with poor eyesight.
These classifications will be used as the buggy labels of the training points, that is, $Y = f^*(\mathrm{blur}(X))$. 
Meanwhile, the original labels for $X$ are retained as the correct labels.
A training image has a buggy label if $Y$ and the original label disagree.
This process gives rise to a total of 133 bugs.
Note that $X$ is always the clear images; the blurred images are only used to generate $Y$.

Step 3. Finally, among the rest, we randomly sample 160 data points, 16 per digit, to form the trusted items $\tilde X$, and use their original labels as trusted labels $\tilde Y$.

Figure~\ref{fig:digit_pr} shows the PR curves that indicate whether each method flags the true bugs, regardless of whether their proposed fix is correct or not.
For this easier task, \HAPPY and INF both excel, but \HAPPY has a slight advantage overall.
Figure~\ref{fig:digit_fix_pr} plots 
the number of flagged bugs vs. the number of correct fixes.
This is a harder task. Nonetheless, \HAPPY still dominates INF and NN, especially in the early stage. Specifically, \HAPPY successfully recovers more than half of the buggy labels (there were 133) within less than 200 attempts.

Table~\ref{tab:digits} visualizes selected buggy training points.
The first row shows the original images,
and the second row shows the wrong label they received from the blurred deepnet.
The next three rows show the actions of \HAPPY, INF, and NN, respectively:
A numerical entry indicates that the debugging method flagged this training image as a bug, and suggested that number as the fixed label.
The entry ``--'' indicates a false negative:  the debugging method missed the bug.

\subsection{Regression Toy Examples}
\begin{figure}[th]
	\centering
	\begin{minipage}[t]{0.45\columnwidth}
		\centering
		\includegraphics[width=0.9\columnwidth]{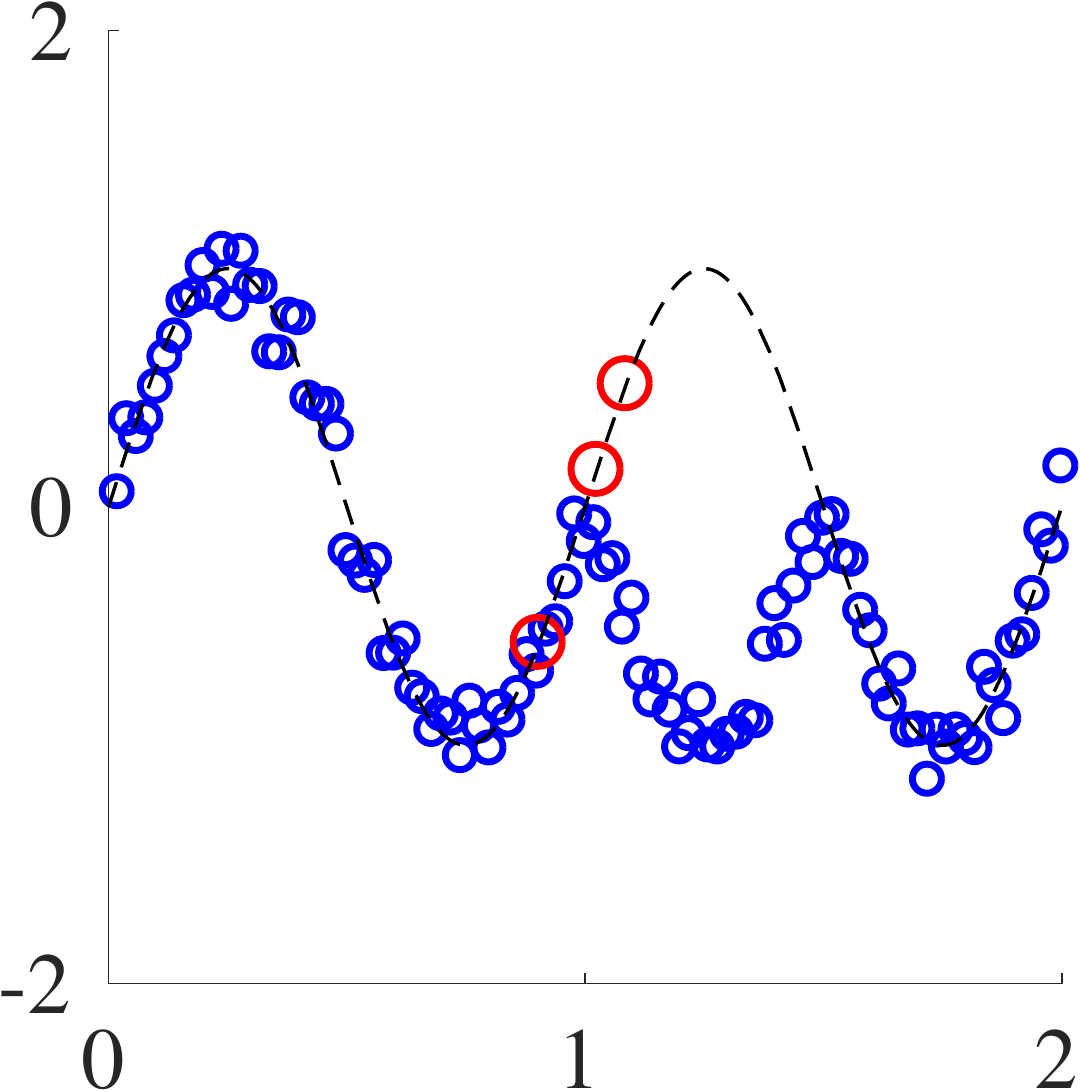} 
		\subcaption{training and trusted data}
		\label{fig:reg_sys_data}
	\end{minipage}%
	~~
	\begin{minipage}[t]{0.45\columnwidth}
		\centering
		\includegraphics[width=0.9\columnwidth]{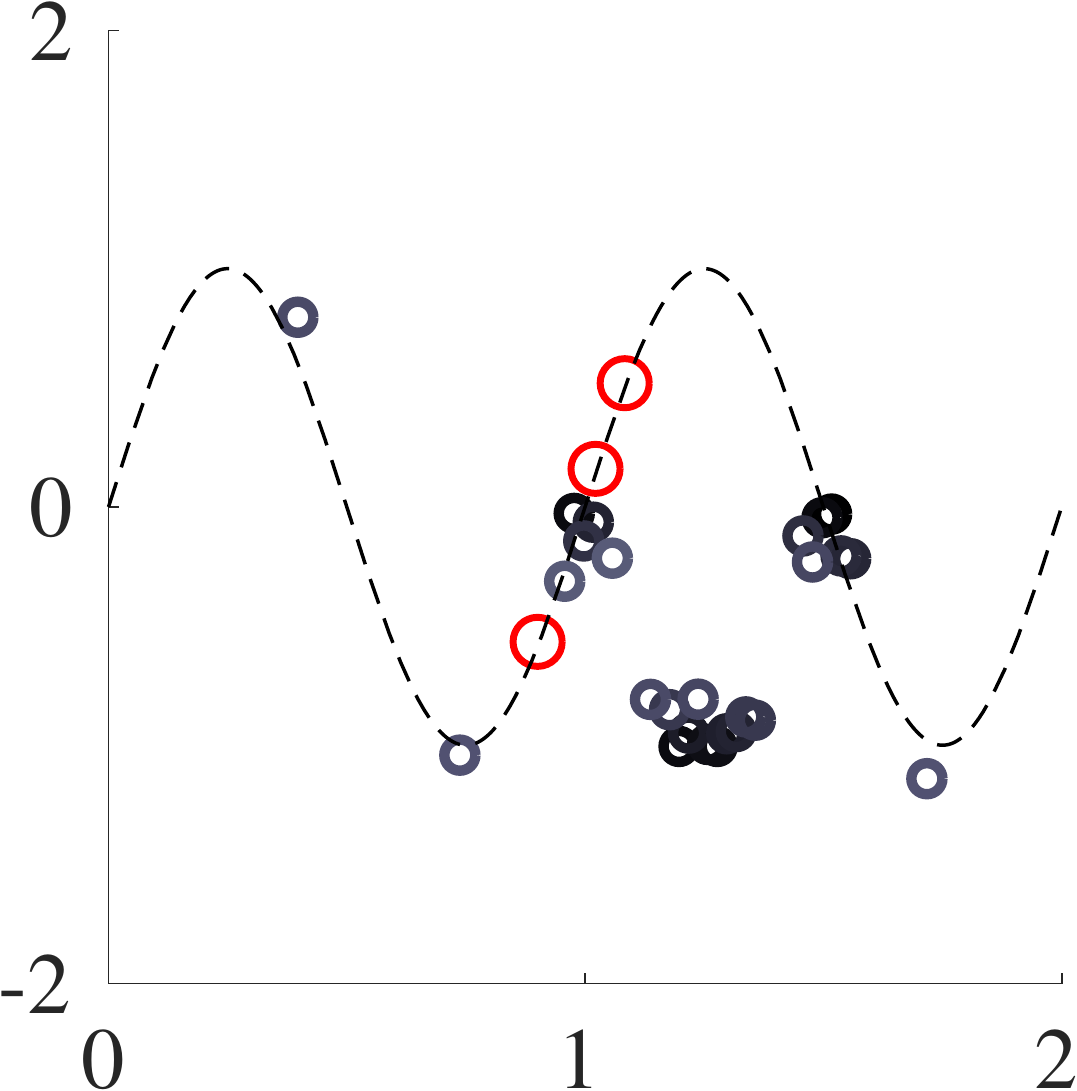}
		\subcaption{\HAPPY flagged bugs}
		\label{fig:reg_sys_duti_flag}
	\end{minipage}
	
	\begin{minipage}[t]{0.45\columnwidth}
		\centering
		\includegraphics[width=0.9\columnwidth]{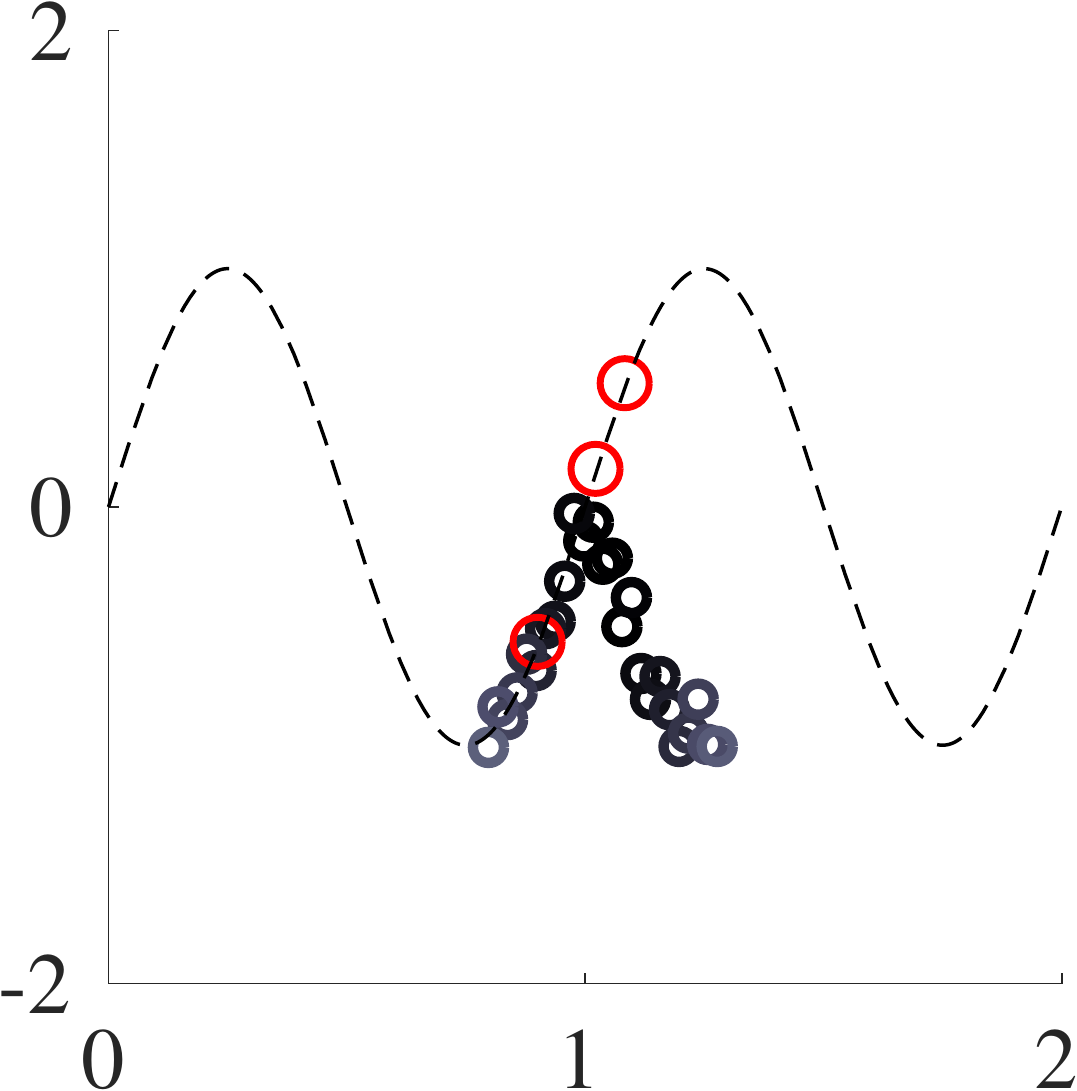} 
		\subcaption{INF flagged bugs}
		\label{fig:reg_sys_inf_flag}
	\end{minipage}%
	~~
	\begin{minipage}[t]{0.45\columnwidth}
		\centering
		\includegraphics[width=0.9\columnwidth]{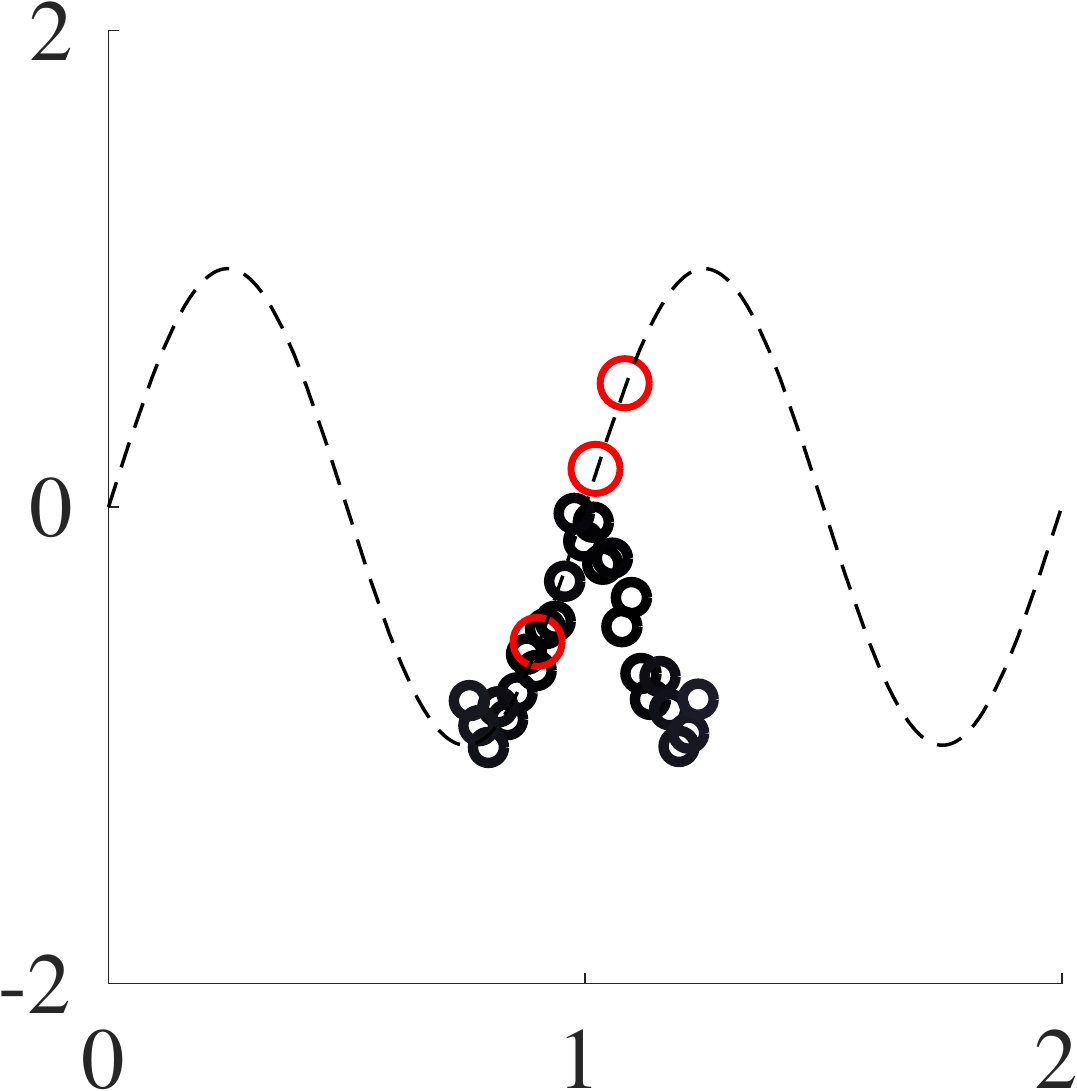}
		\subcaption{NN flagged bugs}
		\label{fig:reg_sys_nn_flag}
	\end{minipage}
	
	\begin{minipage}[t]{0.45\columnwidth}
		\centering
		\includegraphics[width=0.9\columnwidth]{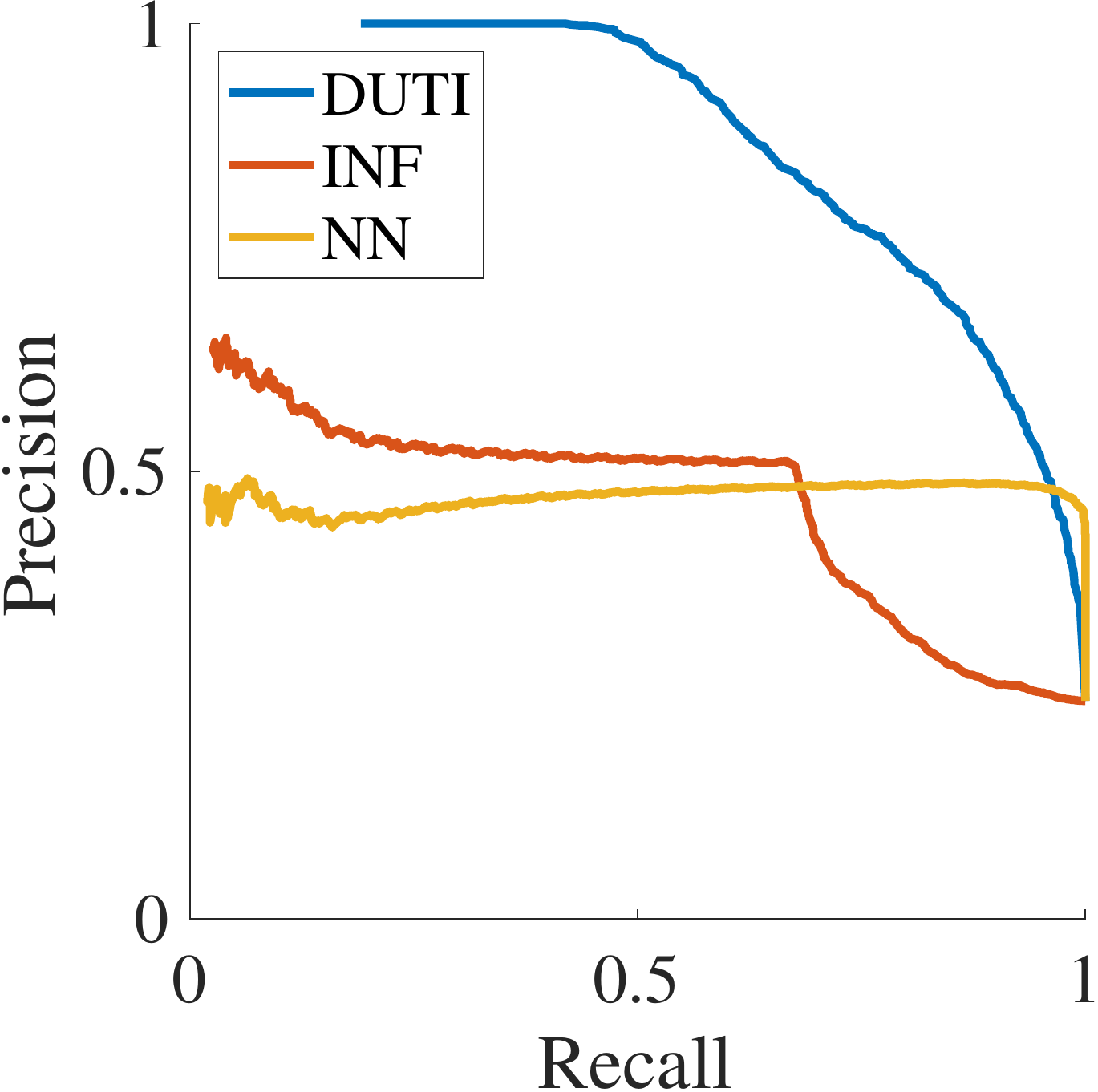} 
		\subcaption{Average PR Curves}
		\label{fig:reg_sys_pr}
	\end{minipage}
	~~
	\begin{minipage}[t]{0.45\columnwidth}
		\centering
		\includegraphics[width=0.9\columnwidth]{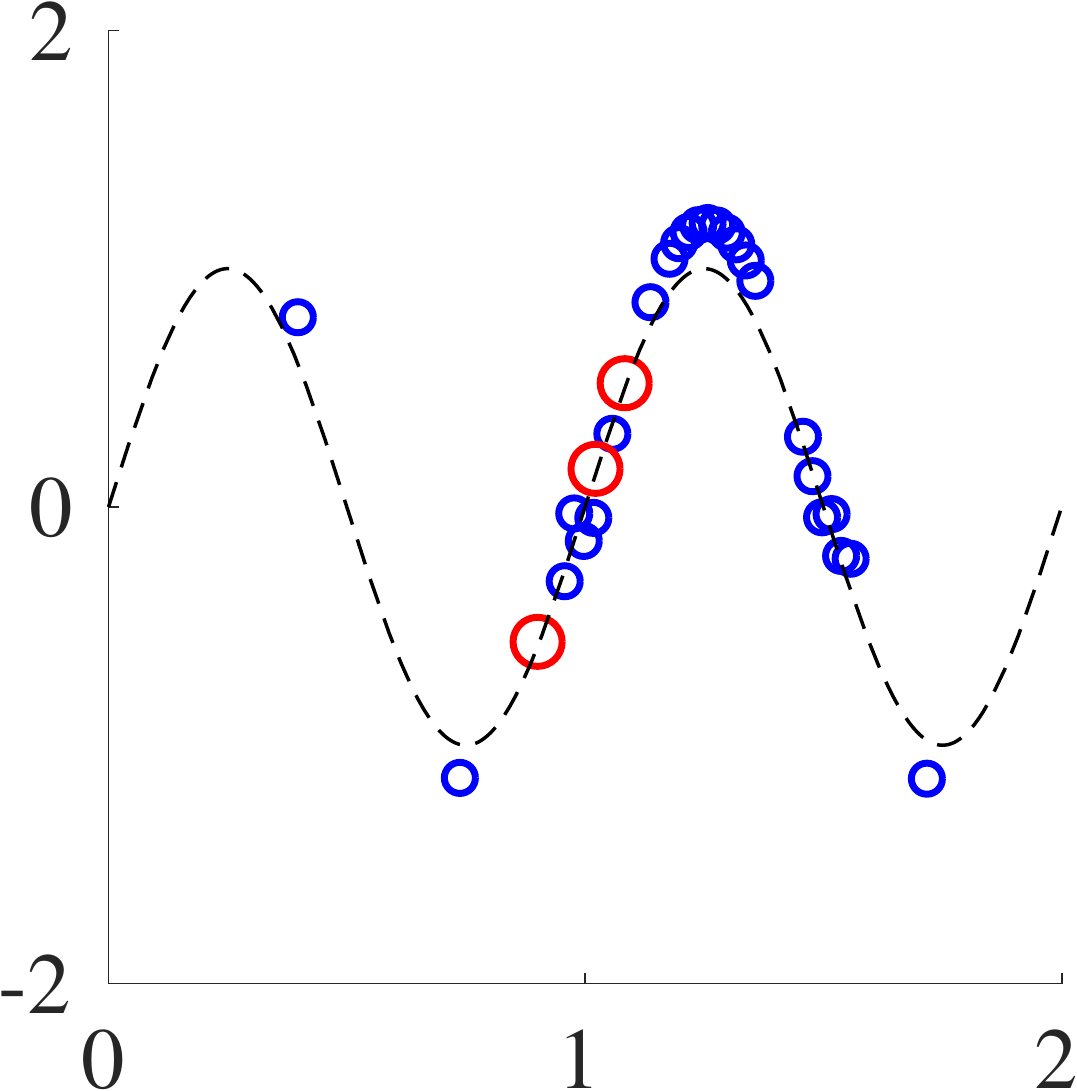}
		\subcaption{\HAPPY suggested fixes}
		\label{fig:reg_sys_duti_fix}
	\end{minipage}%
	\caption{Regression toy data with systematic bugs}
	\label{fig:reg_sys}
\end{figure}
Finally, we demonstrate debugging for regression.
We generate the clean data by uniformly sampling $n=100$ points $x \sim U[0,2]$, and generating their label as
\begin{equation}\label{eq:reg_eqn}
y = \sin(2\pi x) + \epsilon_1\mbox{, where }\epsilon_1\sim N(0,0.1).
\end{equation}
We then manually convert 24 of these points into systematic bugs by flipping the second peak from positive to negative; see Figure~\ref{fig:reg_sys_data}. 
The dashed curve traces out $\sin(2\pi x)$ but is not used in the experiment otherwise.
For trusted items we randomly pick $m=3$ new points near one side of the flipped peak, and generate their $\tilde y$ values using $\tilde y = \sin(2\pi \tilde x)$. 
We intentionally do not fully cover the bug region with the trusted items. 

Figure~\ref{fig:reg_sys}b,c,d plot the top 25 flagged bugs by \HAPPY, INF, and NN, respectively. 
Interestingly, \HAPPY successfully flags bugs throughout the flipped peak. 
In contrast, INF and NN are only able to flag points around the trusted items, and miss half of the bugs.
We repeat the experiment for 100 times, each time with the training data randomly drawn according to~\eqref{eq:reg_eqn}. Figure~\ref{fig:reg_sys}e shows the average PR curves, where \HAPPY clearly dominates INF and NN.
Figure~\ref{fig:reg_sys}f shows the suggested fixes on buggy points (i.e. $y_i+\delta_i$) generated by \HAPPY.
Impressively, the fixes closely trace the sine curve, even though the learner is an RBF kernel regressor and knows nothing about sine function.
This seems to be a smoothing effect of the self-consistent term, which is 
$Predictor(X) = Y'$ in the conceptual formulation~\eqref{eq:trust_constraint} or 
$\frac{1}{n}\sum^{n}_{i=1} \ell(x_i, y_i+\delta_i, \theta )$
in~\eqref{eq:Aregression}.

\section{Related Work}
Debugging has been studied extensively in the programming language community~\cite{shapiro1983algorithmic,ball2002s}. Only recently does the machine learning community start to investigate how to debug machine learning systems, e.g.~\cite{zhu2016debug,bhadra2015}. Our intuition is similar to that of~\cite{zhu2016debug}, who provide closed-form solutions to debugging Ordinary Least Squares. Our work allows one to debug any ERM learners satisfying mild conditions.

More generally, our work falls into the field of interpretable and trustworthy machine learning, where the central question is why a trained model makes certain predictions. One line of work explains the prediction by highlighting the responsible part of the test item. 
For example,~\cite{selvaraju2016grad,smilkov2017smoothgrad,sundararajan2016gradients} identify regions in a test image that have the greatest influence on the classification of that image. In other learning settings where features are less explanatory,~\cite{koh2017understanding} propose to look instead at the most influential training items that affects the prediction on the target test item. Though the influence function is intuitive and easy to compute, we show in experiments that it achieves similar performance to the naive NN baseline, while \HAPPY provides better bug recommendations and has the advantage of producing exact fixes.

Our work is also related to data cleaning studied in data science and statistics. 
Earlier work~\cite{jiang2004editing,zhu2003eliminating,brodley1996improving} applies an ensemble of classifiers to
the training examples, and detects whether the class label
assigned to each example is consistent with the output of
these classifiers. The main problem with this approach
is that the classifiers used to detect mislabeled examples
are themselves constructed from contaminated training items.
Another approach is to design robust statistics with a high break-point against incorrect training examples~\cite{huber2011robust}. 
None of the above-mentioned methods can deal with systematic biases.
Later,~\cite{bhadra2015,valizadegan2007kernel} define a consistency metric and perturb the training labels to maximize the consistency metric. This approach results naturally in a combinatorial optimization problem. In particular, Bhadra and
 Hein's method is able to incorporate expert verified data as hard constraints in their optimization formulation. However, their methods requires information on the exact number of bugs which is often not available. 
\HAPPY does not request such information.

 Along the line of incorporating outside information in training set debugging,~\cite{ghosh2016trusted} requires the learned model to satisfy certain Probabilistic Computation Tree Logic (PCTL).
That is, the bugs are revealed by the learned model violating logical constraints rather than by the trusted items.
This approach complements \HAPPY and the two can potentially be combined in the future.
We note that in complex machine learning applications it can be more difficult for experts to encode domain knowledge in rigorous logical statements, whereas providing verified trusted items is often easier. 

Our work is partly inspired by the need to identify historical unfairness in a training set.
\HAPPY thus joins recent work~\cite{hardt2016equality,corbett2017algorithmic,zemel2013learning,feldman2015certifying} on algorithmic fairness.
In our experiments, we demonstrated how \HAPPY can be used to identify historical discrimination, with the hope that such identification will improve fairness of machine learning systems.

\section{Limitations of \HAPPY and Future Work}

This paper contributed to trustworthy machine learning by proposing a novel training set debugging algorithm \HAPPY.
As the experiments demonstrated, \HAPPY was able to detect and fix many bugs in diverse data sets.
Like any method, \HAPPY has its limitations.  We discuss three major ones.
\begin{figure}[th]
	\begin{center}
		\begin{tabular}{cc}
			\includegraphics[width=.4\columnwidth]{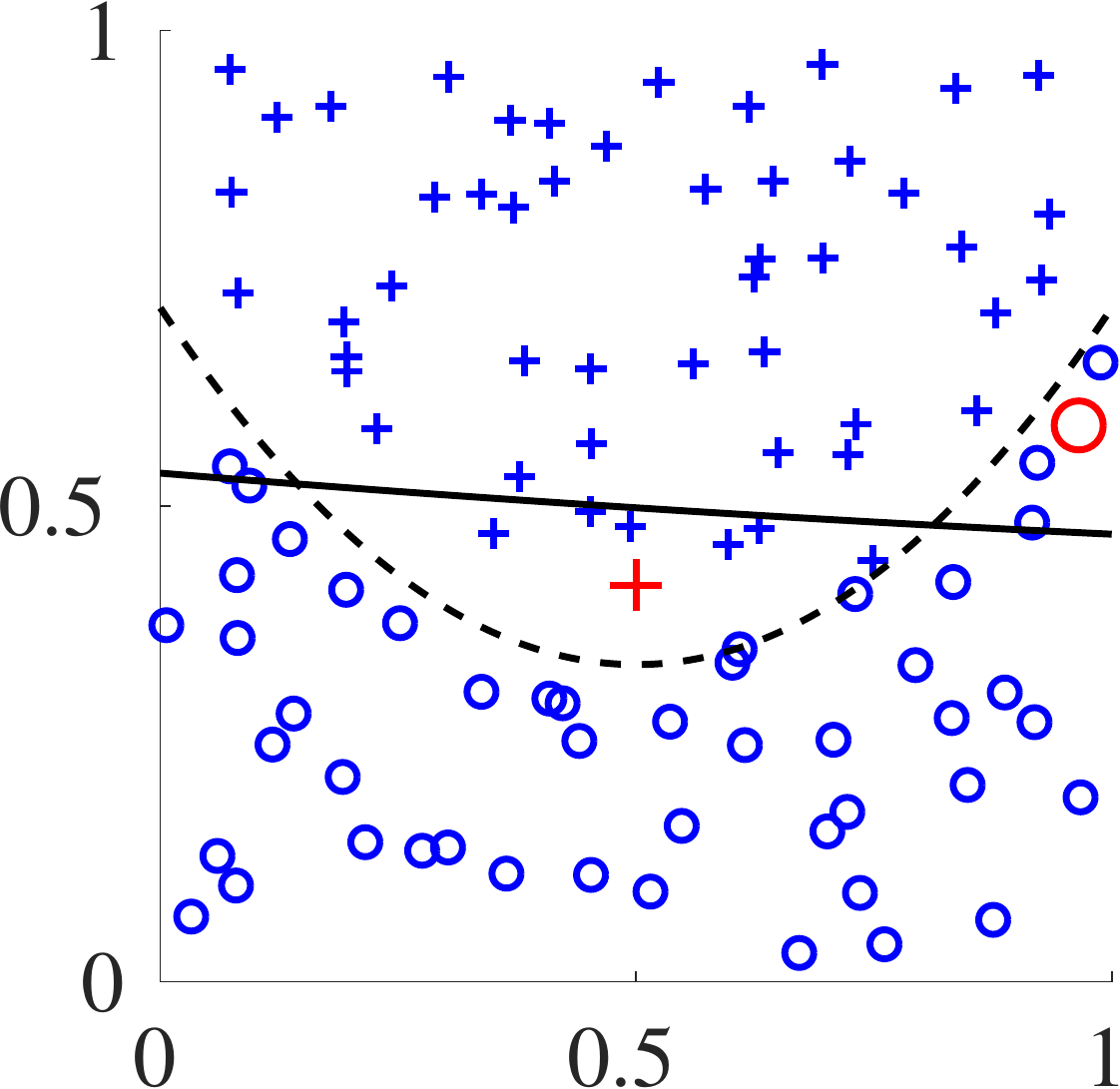} &
			\includegraphics[width=.4\columnwidth]{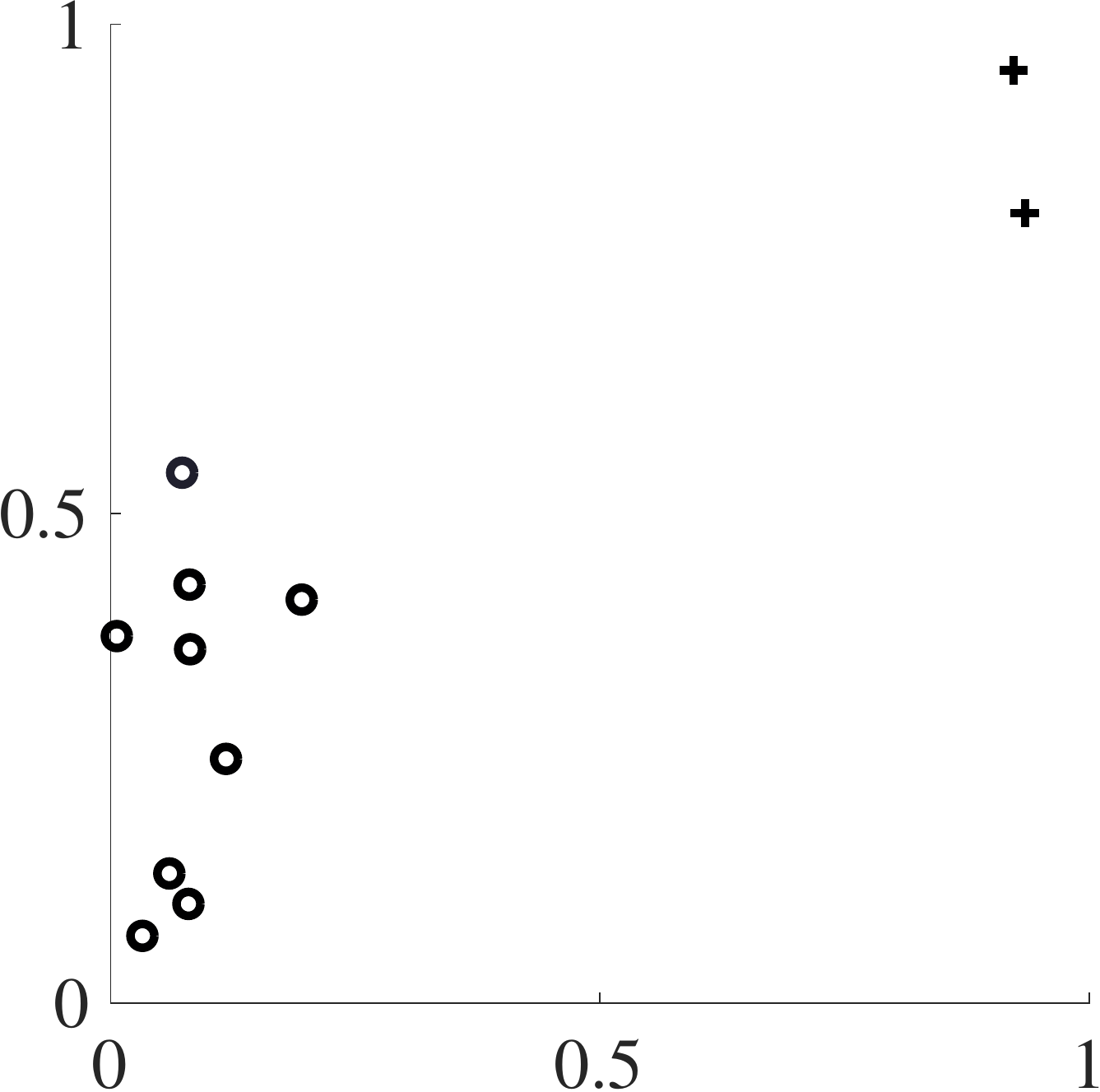} \\
			(a) underfitting & (b) \HAPPY flagged ``bugs'' \\
			\includegraphics[width=.4\columnwidth]{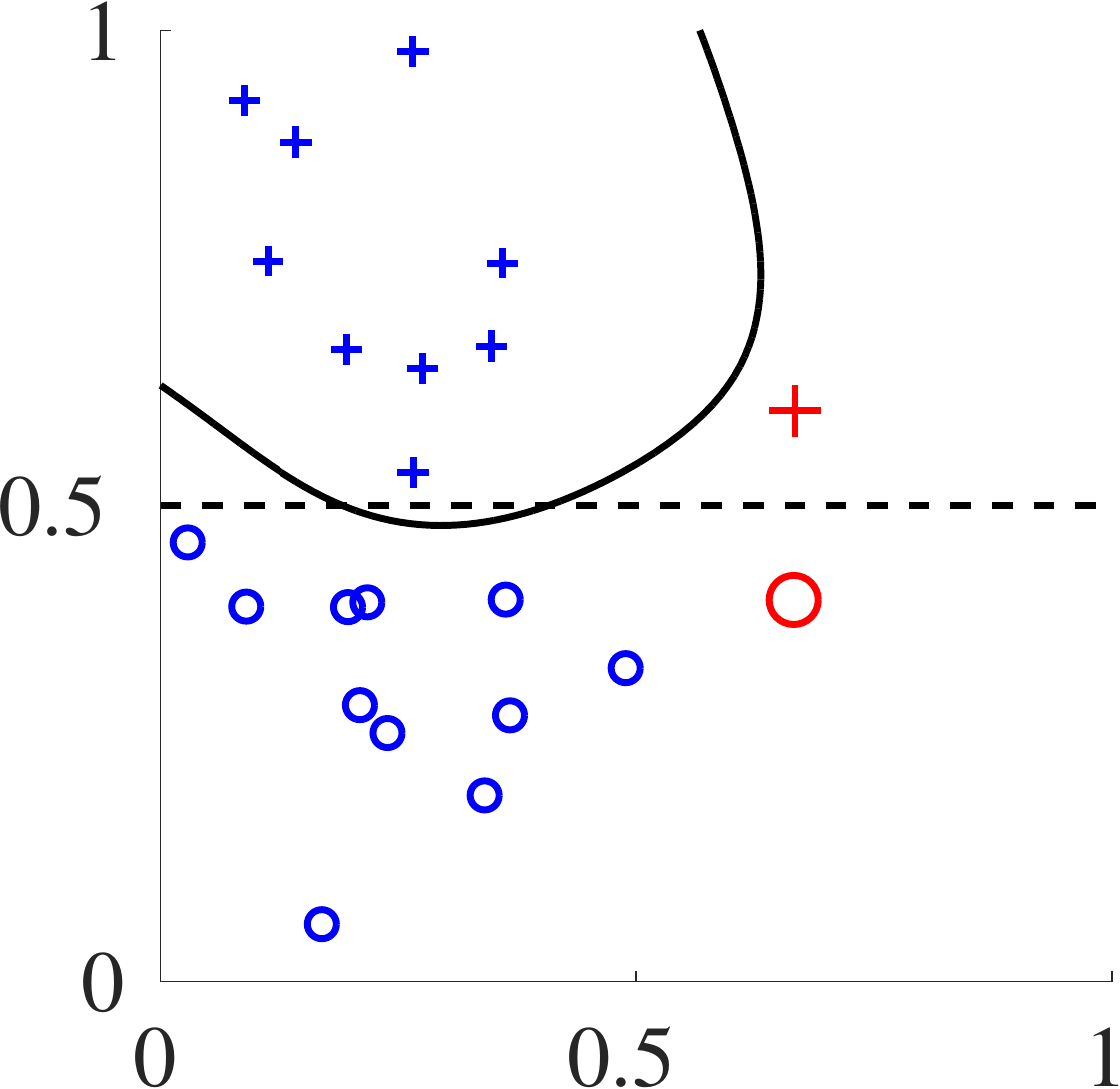} &
			\includegraphics[width=.4\columnwidth]{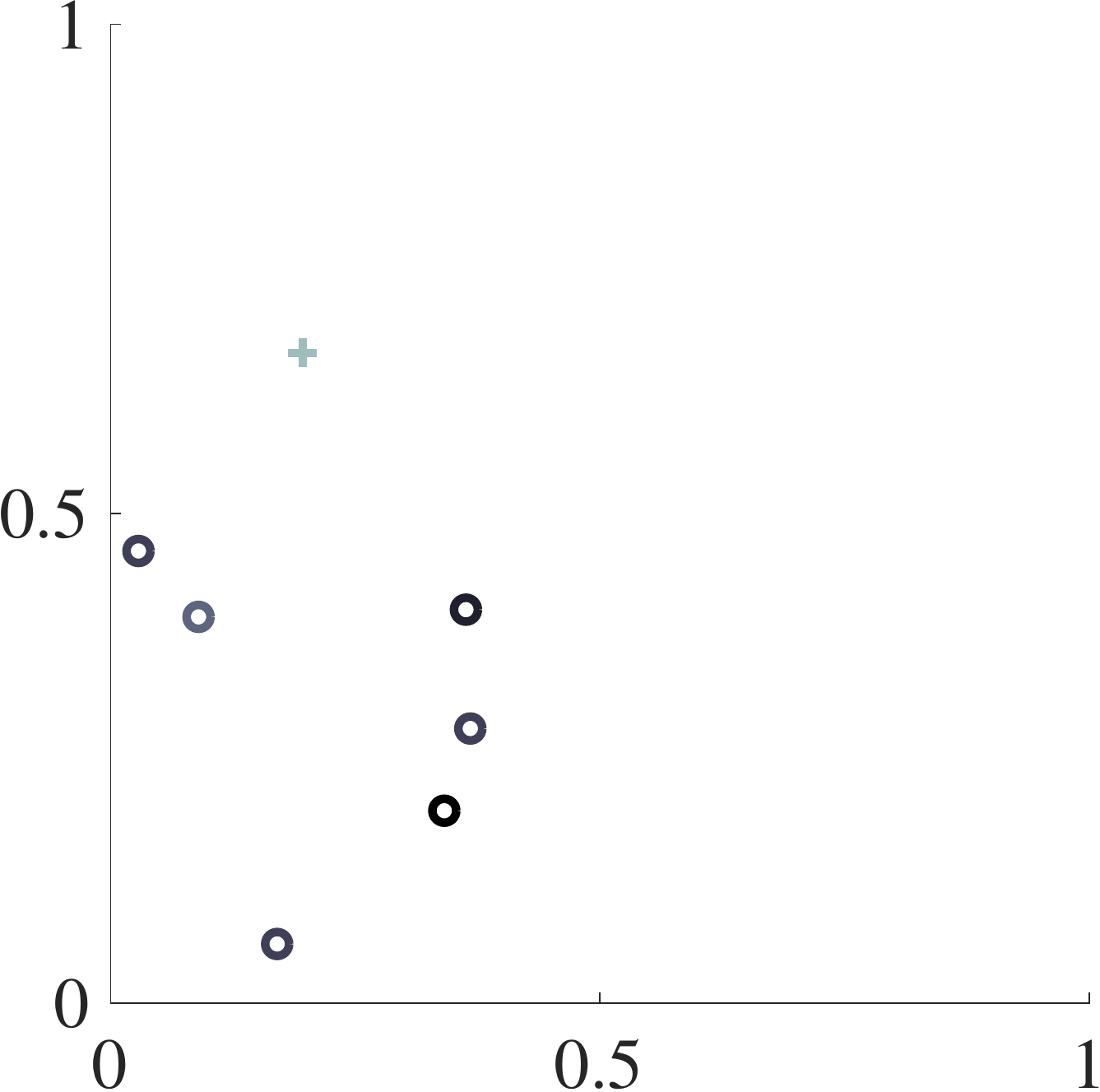} \\
			(c) poor extrapolation & (d) \HAPPY flagged ``bugs'' \\
			\includegraphics[width=.4\columnwidth]{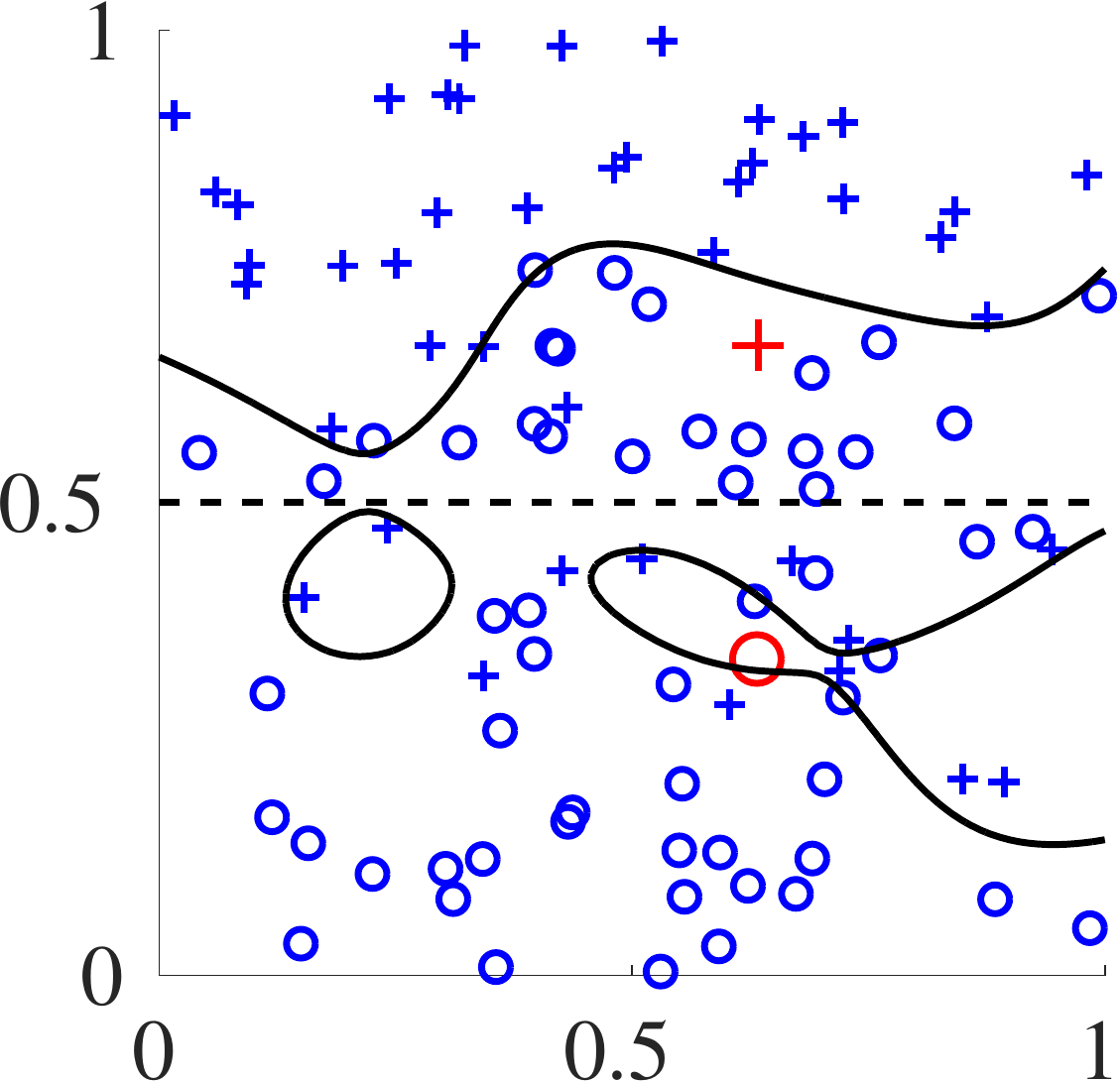} &
			\includegraphics[width=.4\columnwidth]{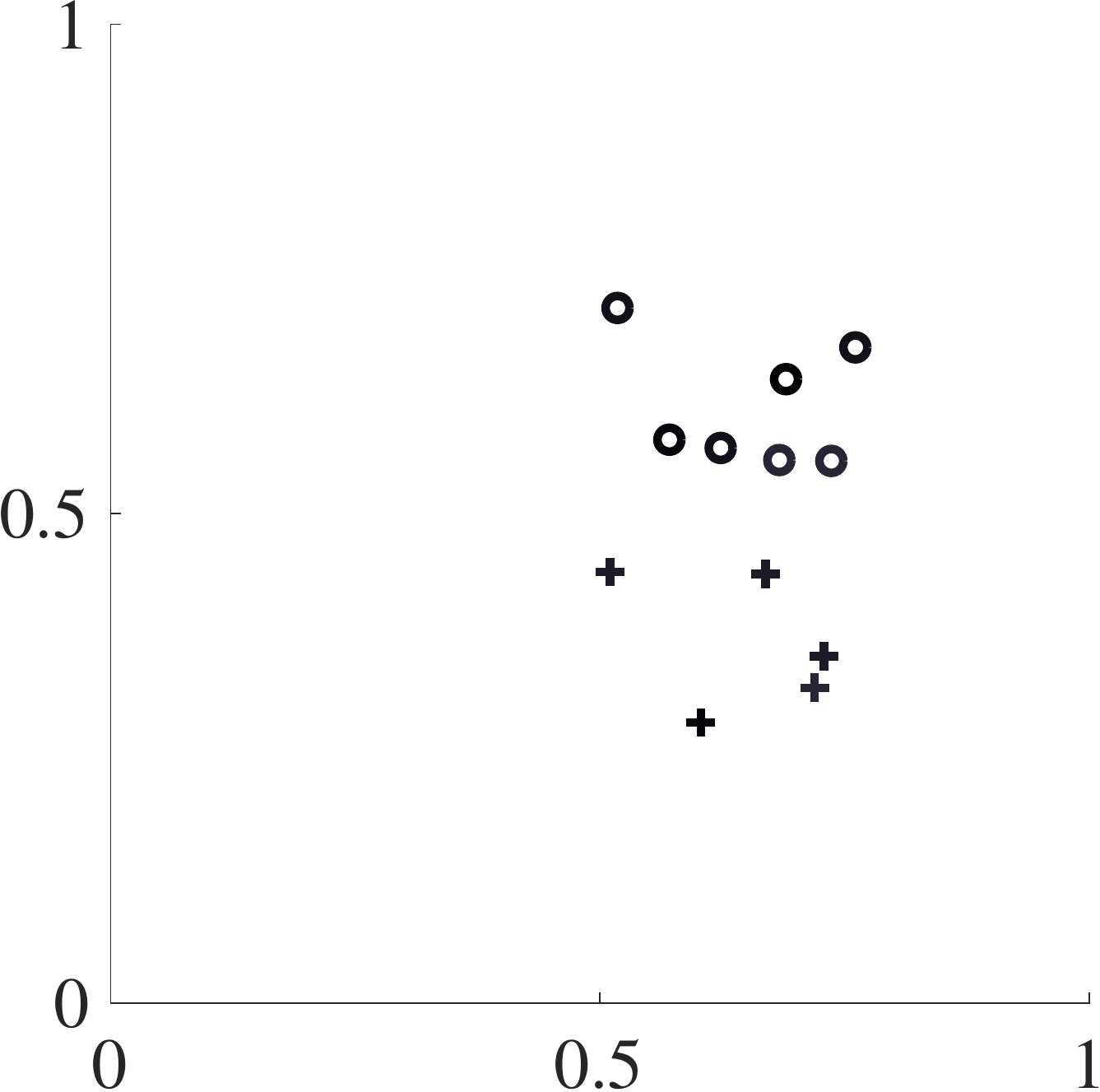} \\
			(e) high Bayes error & (f) \HAPPY flagged ``bugs''  
		\end{tabular}
	\end{center}
	\caption{Trusted item violations not caused by label bugs}
	\label{fig:other}
\end{figure}

\textbf{Applicability:} While a violated trusted item often indicates training set label bugs, it is not always the case.  A domain expert needs to bear in mind other reasons that violate a trusted item while nothing is wrong with the training labels. 
Figure~\ref{fig:other}(a,c,e) presents three common cases.
In all cases, the dashed line is the true decision boundary.
The blue points represent the training set, which obey (stochastically in the third case) the true boundary.
Therefore, there is no bug in the training labels \emph{per se}.
The red points are the trusted items, which also obey the true boundary.
However, the solid curve is the boundary learned from the training set.
In all three cases, some trusted items are indeed violated by the learned boundary:
In (a), the true boundary is nonlinear but the hypothesis space is linear, resulting in underfitting and thus violating both trusted items.
In (c), the trusted items are in a region of the feature space poorly covered by training data, thus unreliable extrapolation.  This happens in covariate shift, for example.
In (e), the underlying conditional distribution $P(Y\mid X)$ has high Bayes error (e.g. close to 0.5 near the true boundary), and the hypothesis space is too rich and ill-regularized, resulting in overfitting.

In all three cases, it is inappropriate to apply \HAPPY in the first place.
In fact, blindly running \HAPPY will result in the ``flagged bugs'' in (b, d, f), respectively, and none are true bugs.
Conversely, after the domain expert verifies that none of \HAPPY flagged items are bugs, she should suspect some of the above reasons in the machine learning pipeline.

\textbf{Theoretical Guarantee:} The trusted items need to be informative for \HAPPY to work.
For example, in Figure~\ref{fig:hp}(a) if the two trusted items were ``hired'' at $(1,1)$ and ``not hired'' at $(1,0)$, they would still be correct but would not have revealed the systematic bias in the data.
In our real data experiments trusted items are i.i.d samples, but DUTI does not require this.
Future work should study theoretical guarantees of (potentially non-$iid$) trusted items on debugging.

\textbf{Scalability:} The current implementation of \HAPPY has limited speed and scalability. 
At each step of optimization~\eqref{eq:Aregression} or~\eqref{eq:Aclassification}, it has to compute $\theta(\delta)$ which is equivalent to training the learner.
Even with smart initialization, this repeated learning subroutine still takes the majority of time. 
For large data sets, one iteration can take minutes.
\HAPPY currently can handle training set size $n$ in the thousands.
Future work is needed to yield a faster algorithm and implementation.

To conclude, in this work we designed DUTI, an efficient algorithm that helps the users to identify and correct training set bugs on any ERM learner, with the help of verified trusted items. Empirical experiments demonstrated that DUTI is able to tackle different types of realistic systematic bugs and outperforms other related methods. Future work will be dedicated to building a general theory of debugging and improving scalability through smart optimization.

\textbf{Acknowledgment}.
This work is supported by
NSF Awards 1704117, 1623605, 1561512, 1545481, 1447449, 1634597, and 1740707; AFOSR Award FA9550-13-1-0138; and Subcontract 3F-30222 from Argonne National Laboratory.

%

\bibliographystyle{aaai}
\bibliography{nllb}

\end{document}